\documentclass[10pt,twocolumn,letterpaper]{article}

\usepackage[pagenumbers]{cvpr}              

\definecolor{cvprblue}{rgb}{0.21,0.49,0.74}
\usepackage[pagebackref,breaklinks,colorlinks,allcolors=cvprblue]{hyperref}

\usepackage{wrapfig}
\usepackage{algorithm}
\usepackage[noend]{algorithmic}
\usepackage{amsmath}
\usepackage{multicol}
\usepackage{tabularx}
\usepackage{svg}
\usepackage{adjustbox}
\usepackage{tikz}

\def\paperID{1482} 
\def\confName{CVPR}
\def\confYear{2025}

\title{Multidimensional Byte Pair Encoding:\\Shortened Sequences for Improved Visual Data Generation}

\author{
Tim Elsner, Paula Usinger, Julius Nehring-Wirxel, Gregor Kobsik,\\Victor Czech, Yanjiang He, Isaak Lim, Leif Kobbelt\\~\\
Visual Computing Institute, RWTH Aachen University\\
Aachen, Germany}

\begin{document}
\maketitle

\begin{abstract}
In language processing, transformers benefit greatly from text being condensed. This is achieved through a larger vocabulary that captures word fragments instead of plain characters. This is often done with Byte Pair Encoding. In the context of images, tokenisation of visual data is usually limited to regular grids obtained from quantisation methods, without global content awareness.\\
Our work improves tokenisation of visual data by bringing Byte Pair Encoding from 1D to multiple dimensions, as a complementary add-on to existing compression. We achieve this through counting constellations of token pairs and replacing the most frequent token pair with a newly introduced token. The multidimensionality only increases the computation time by a factor of 2 for images, making it applicable even to large datasets like ImageNet within minutes on consumer hardware. This is a lossless preprocessing step.\\
Our evaluation shows improved training and inference performance of transformers on visual data achieved by compressing frequent constellations of tokens: The resulting sequences are shorter, with more uniformly distributed information content, e.g. condensing empty regions in an image into single tokens. As our experiments show, these condensed sequences are easier to process. We additionally introduce a strategy to amplify this compression further by clustering the vocabulary.
\end{abstract}
\section{Introduction}
\begin{figure}[h]
    \centering
    \includegraphics[width=\columnwidth]{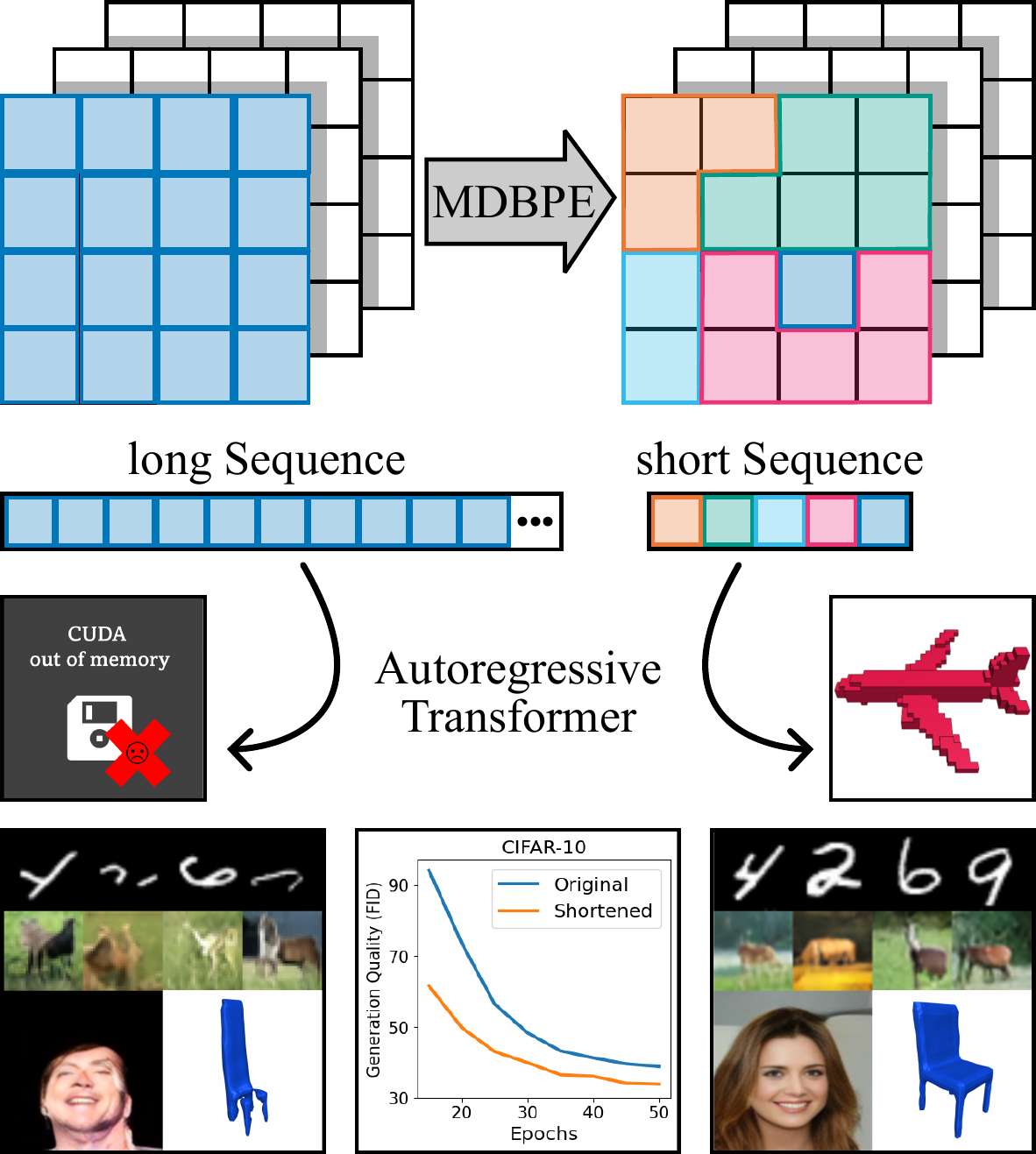}
    \caption{Our algorithm compresses visual data in order to make tasks like generation more efficient: Shorter sequences, even if they are from a larger vocabulary, are easier to handle for deep learning architectures like transformers. The images show representative examples after the same training time, with training on shortened sequences (right) producing better results faster.}
    \label{fig:teaser}
\end{figure}

Much of the recent success and widespread application of deep learning has come with the understanding that generation can form a building block, a \textit{foundation model}, for further tasks. These models then can be tuned to various tasks for practical applications, from chat assistants \cite{llama3, gpt4, openassistant} to image \cite{dai2023emu, edibert, cheong2023upgptuniversaldiffusionmodel} or video editing \cite{meta_movie, feng2024ccedit,shi2024bivdifftrainingfreeframeworkgeneralpurpose}.\\
For natural language processing (NLP), condensation from long sequences with small vocabulary into shorter sequences with larger vocabulary significantly improves the training of models. This approach not only increases efficiency by encoding more information per token, resulting in faster and more effective learning, but also plays well with the quadratic scaling of the underlying attention mechanism \cite{base_attention} that has been driving many recent improvements \cite{attention}. In practice, it is easier to assemble sentences from word fragments rather than individual characters. One widespread strategy to obtain larger vocabularies is Byte-Pair Encoding (BPE) \cite{bpe, bpe2}. By merging frequent pairs of tokens - initially single letters - BPE forms compact, reusable word fragments. This approach offers two main advantages: It is \textit{lossless}, meaning it preserves information while enhancing efficiency, and it is \textit{adaptive}, balancing what is best to compress given a certain vocabulary budget.\\
\\
For visual data like images, condensation often happens in a content agnostic grid format, \eg through autoencoders \cite{vqvae, vqvae2, first_ae, hinton2006reducing}, which does not address uneven information density: A patch of uniform colours in the background is represented through the same amount of tokens as a patch containing complex patterns in the foreground. While global information mechanisms such as attention or U-Net \cite{unet} architectures can enhance learned representations by distributing information more evenly, they do not fundamentally resolve the issue: The resulting compression is not \textit{adaptive} to the content.\\
Adaptive approaches like Byte Pair Encoding on images that are reshaped into 1D sequences only apply compression along one axis. Therefore, we extend this form of compression to true multidimensionality to yield similar benefits for visual data as for BPE on text.\\
Our multidimensional approach retains the strengths of traditional BPE, \ie it is also lossless and adaptive, while offering more flexibility in the choice of pairs and can be integrated as an add-on into various discrete setups seamlessly \cite{vqvae, vqvae2, vqgan, PLBPE}. We see applications not only for faster generation, but also \eg in multimodal LLMs, where fewer tokens to represent information accurately is of the essence. \\
This work is therefore centred around a simple hypothesis: \textbf{Shorter sequences from visual data are better suited for processing with neural networks}. Our contributions are as follows:
\begin{itemize}
    \item Motivated by 1D token compression from NLP, we provide a preprocessing step to compress visual data.
    \item We show how shortened sequences with varying lengths can be better processed by transformers.
    \item To further improve effectiveness, we show a lossy variant to improve compression rates for discrete tokens.
    \item We evaluate the improved performance of the resulting compressed sequences on generation tasks.
\end{itemize}
Our work leads to earlier convergence at better final scores, while enabling \eg large sizes of voxel grids that would otherwise not fit into the memory in the first place. It further aligns well with the recent trend of including multimodal data into LLMs \cite{llama3, gpt4}.\\
We provide the complete codebase for our approach as easily accessible Jupyter Notebooks, along with a faster C++ implementation for the compression algorithm at the core\footnote{We provide a small MNIST demo that can be run within minutes here: \href{https://github.com/DaiDaiLoh/MDBPE_TF}{https://github.com/DaiDaiLoh/MDBPE\_TF}~~~There, we will eventually release the full version including the faster C++ code.}.
\section{Related Work}
As closely related fields, we discuss traditional compression and token-based learning (\cref{rw:token_based}). We also discuss compression in the context of learning and generating visual data from discrete representations (\cref{rw:neural}).

\subsection{Compression for Token-Based Learning}\label{rw:token_based}
While initial tokenisers were based on abstracting words through tokens \cite{maxmatch1, maxmatch2}, more recent schemes have become adaptive in the sense of allowing subword tokenisation to replace character sequences instead of words. As the baseline tokenisation scheme used for many recent LLMs \cite{gpt4, llama3}, Byte Pair Encoding (BPE) \cite{bpe, bpe2} computes the most frequent combination of elements to produce new tokens. Other approaches like the work of Song \etal \cite{wordpiece} focus on improving compression speed, or finding tokenisation in the context of sentences \cite{kudo2018sentencepiecesimplelanguageindependent}. Recent approaches \cite{tfree, wang2024mambabyte} move away from assigning a single sequence part to a single token and instead combine different base elements. Razzhigaev \etal \cite{PLBPE} apply Byte Pair Encoding directly to quantised images, but only for images turned into 1D sequences.
\subsection{Neural Compression and Generative Models}\label{rw:neural}
While autoencoders are generally built to compress data \cite{first_ae, hinton2006reducing} through a narrowing network architecture, autoregressive approaches like transformers \cite{transformer} typically require a finite and discrete set of tokens to learn a probability distribution over the next token that can be sampled from.
To obtain such quantised and compressed images of high quality, van der Oord \etal introduced the VQ-VAE family \cite{vqvae, vqvae2}, applying vector quantisation through rounding latent vectors to a set of codebook vectors. While some models deviate \eg by de-noising discrete tokens \cite{paella}, most generative models then employ a transformer architecture \cite{dalle,vec_gen,vq_video} on the discretised representation.\\
To improve image quality of VQ-VAEs, approaches either improve compression efficiency, \eg through avoiding unused codebook entries \cite{wu2020vector, lancucki2020robust, cvqvae, vqwae, svqvae, qgvae}, or add an additional discriminator to "reinvent" high-frequent details, also known as VQGAN \cite{vqgan, vqgan_alt}, which still forms the basis for recent work in image and video generation \cite{meta_movie, zhu2023designing, cao2023efficient}.\\
Other methods focus on improving compression by moving away from rigid grids of compact "super pixels", \eg further compressing grids by transformers \cite{koner2024lookupvit}, or by avoiding a representation based on a (non-adaptive) grid layout entirely \cite{qgvae, titok}. Alternatively, some methods focus on introducing content adaptivity for images \cite{dqvae} and geometry \cite{octree}. Often, these approaches are more difficult to implement than autoregressive generation.\\
Aside from autoregressive models, plenty of approaches from the diffusion \cite{diffusion} and flow matching \cite{lipman2022flow} families learn to map an input distribution to a target distribution (\ie a dataset of images), while generation can also be based on min-max games (GANs) \cite{gan, sgan}. These can provide high image quality, but they are often more cumbersome to work with than autoregressive generation and hence not the focus of our work.
\section{Preliminaries: Byte Pair Encoding}\label{rw:bpe}
In 1D token sequences, traditional Byte Pair Encoding iteratively reduces the sequence length by counting the occurrences of adjacent token pairs and replacing the most frequent pair with a new token. This process is repeated until the sequence reaches the desired length in number of tokens (\textit{vocabulary size}). For example, LLMs from the Llama family use a vocabulary of 128,256 tokens \cite{llama3}.
We demonstrate BPE with a small example:\\
\begin{tabularx}{\columnwidth}{
    | l 
    >{\raggedright\arraybackslash}X | }
\hline
 \textbf{Sequence:} & \texttt{AAAAABBABABB}\\ 
 \textbf{Count:} &  $4 \times$ \texttt{AA}, $3 \times$ \texttt{AB}, $3 \times$ \texttt{BA}, $2 \times$ \texttt{BB}\\  
 \textbf{Replace:} & \texttt{AA} with \texttt{C}\\
 \textbf{Sequence:} & \texttt{CCABBABABB}\\
 \textbf{Count:} & $1 \times$ \texttt{CC}, $1 \times$ \texttt{CA}, $3 \times$ \texttt{AB}, $2 \times$ \texttt{BB}, $2 \times$ \texttt{BA}\\
 \textbf{Replace:} & \texttt{AB} with \texttt{D}  \\
 \textbf{Sequence:} & \texttt{CCDBDDB}\\
 \textbf{Count:} & $1 \times$ \texttt{CC}, $1 \times$ \texttt{CD}, $2 \times$ \texttt{DB}, $1 \times$ \texttt{BD}, $1 \times$ \texttt{DD}\\
 \textbf{Replace:} & \texttt{DB} with \texttt{E}  \\
 \textbf{Sequence:} & \texttt{CCEDE}\\
 \hline
\end{tabularx}
\\

\section{Multidimensional Sequence Compression}
\begin{wrapfigure}{r}{0.22\textwidth}
  \centering
  \vspace{-0.3cm}
  \hspace{-0.5cm}
  \includegraphics[width=0.48\columnwidth]{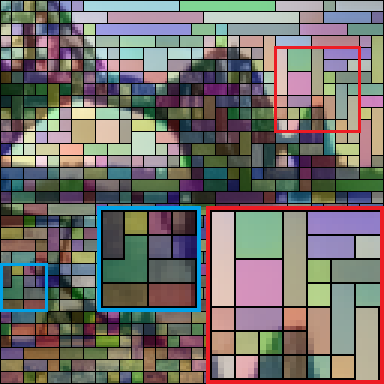}
\end{wrapfigure}
As demonstrated for standard 1D sequences \cite{llama3, gpt4} and linearised 2D sequences of visual data like images \cite{PLBPE}, Byte Pair Encoding \cite{bpe, bpe2} is highly beneficial when processing tokens, especially in generative settings. Inspired by the idea of counting and then replacing tokens with a pairwise mask for neighbouring tokens, we generalise the idea from $1$-dimensional to $n$-dimensional, referred to as \textit{Multidimensional Byte Pair Encoding (MDBPE)}. This is a pre-processing step for the input data, just as regular BPE is for text.\\
Our algorithm allows a better compression for high dimensional data by using more directions to identify pairs of tokens and supports combining patterns into
non-convex tokens for stronger compression.
\subsection{Multidimensional Byte Pair Encoding}
\begin{figure*}[ht]
    \centering
    \begin{minipage}{0.31\textwidth}
        \centering
        \includegraphics[width=0.8\textwidth]{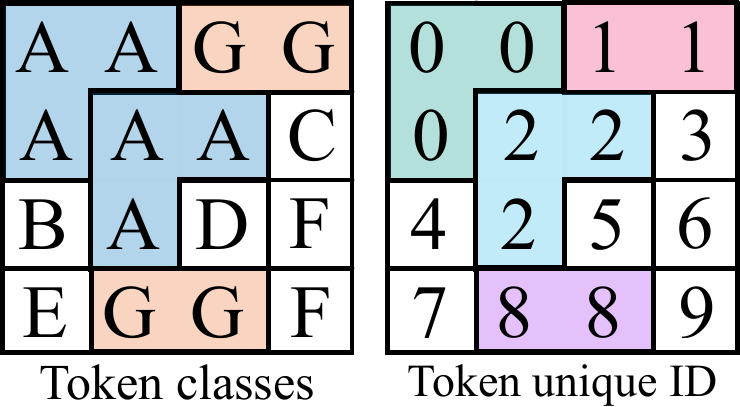}
        \caption{Token classes and unique token ID. The resulting sequence would be \textbf{AGACBDFEGF} (compressed from length $16$ to length $10$).}
        \label{fig:token_unique_class}
    \end{minipage}
    \hfill
    \begin{minipage}{0.35\textwidth}
        \centering
        \includegraphics[width=\textwidth]{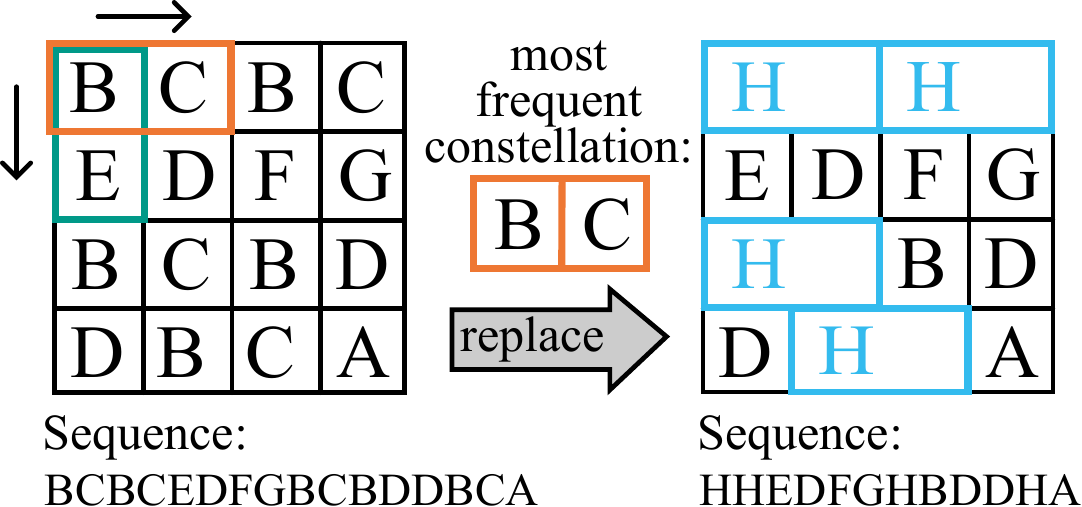}
        \caption{We replace token constellations occurring frequently by sliding a pairwise mask over each dimension.}
        \label{fig:replacements}
    \end{minipage}
    \hfill
    \begin{minipage}{0.29\textwidth}
        \centering
        \includegraphics[width=0.5\columnwidth]{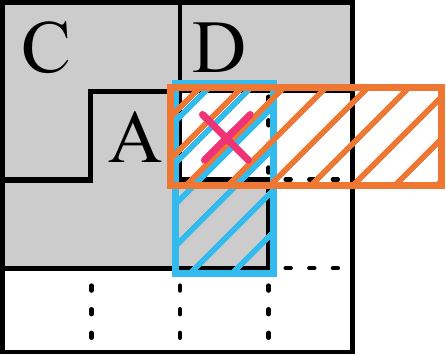}
        \caption{For predicting the next token at a position (red cross), we can reroll tokens that exceed the boundary (orange) or overlap with existing tokens (blue).}
        \label{fig:not_generateable}
    \end{minipage}
\end{figure*}
While we illustrate our approach using images, the same principle is applied across various types of data, as we show with $3$D voxel grids \cref{eval:geometry}. 
For better readability, we further refer to the cells in a data grid as pixels.\\
\textbf{We alternate between two steps:} We first \textbf{count} neighbouring tokens not only in the 1D direction $x=(1, 0)$, but also in direction $y=(0, 1)$ (horizontally and vertically). Diagonal connections provided minimal compression improvements during our experiments and were hence omitted for simplicity. Then, we \textbf{replace} the most frequent token combination by a new token (\textit{extra token}). We alternate the two steps until we reach the desired vocabulary size.
\begin{wrapfigure}{r}{0.25\textwidth}
  \centering
  \vspace{-0.3cm}
  \hspace{-0.6cm}
  \includegraphics[width=0.5\columnwidth]{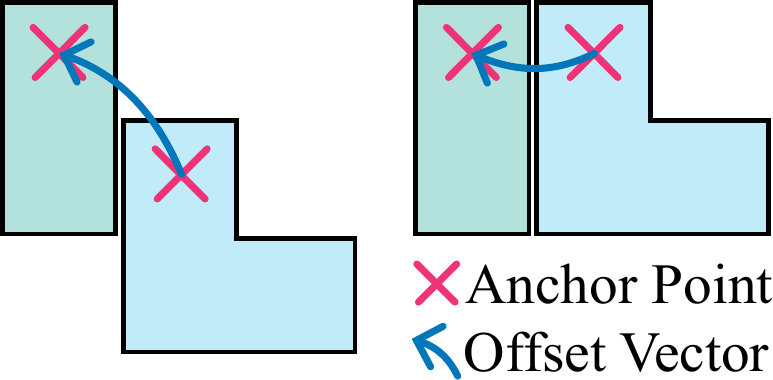}
  \vspace{-0.3cm}
\end{wrapfigure}
As we can have various \textit{constellations} of the same pair of token classes, it becomes necessary to distinguish these for counting, \ie count not pairs of tokens, but tuples of tokens and their alignment (see inset). To solve this, we introduce \textit{anchor points}, which act as a canonical reference point, chosen as the leftmost pixel in the uppermost row. A constellation is determined by their tokens $p$ and its neighbour $n$ and their alignment, \ie characterised by vector $v_{pn} \in \mathbb{Z}^2$ between their anchor points. In the sequence, each token will later be represented at its anchor point. For an uncondensed sequence ($0$ newly introduced extra tokens), this means each pixel is an anchor point.

\begin{algorithm}
\caption{Multidimensional Byte Pair Encoding}\label{alg:MDBPE}
\begin{algorithmic}[1]
\STATE $class \gets \text{class of the token a pixel belongs to}$
\STATE $id \gets \text{unique ID of the token a pixel belongs to}$
\WHILE{$|\text{vocabulary}| < K$}
    \STATE $count \leftarrow 0$
    \FORALL{images}
        \STATE $used \leftarrow \{\}$ 
        \FORALL{grid points $g$ in the image}
            \STATE $p \leftarrow$ token in $g$
            \FORALL{neighbours $n$ of $p$}
                \IF{$p.id = n.id$ or \\ $(p.id,\ n.id) \in used$}
                    \STATE \textbf{continue}
                \ENDIF
                \STATE Add $(p.id,\ n.id)$ to $used$
                \STATE $v_{pn} \leftarrow p.anchor - n.anchor$
                
                \STATE $key \leftarrow (p.class, n.class,\ v_{pn})$
                \STATE $count[key]~+= 1$
            \ENDFOR
        \ENDFOR
    \ENDFOR
    \STATE $constellation \leftarrow \arg\max_{k} \ count[k]$
    \FORALL{images} 
        \FORALL{grid points $g$ in the image}
            \STATE $p \leftarrow$ token in $g$
            \FORALL{neighbours $n$ of $p$}
                \STATE $v_{pn} \leftarrow p.anchor - n.anchor$
                
                \IF{$constellation = (p.class,\ n.class,\ v_{pn})$}
                    \STATE $ind_p \leftarrow \{i \mid i.id = p.id\}$
                    \STATE $ind_n \leftarrow \{i \mid i.id = n.id\}$
                    
                    \STATE $id[ind_n] \leftarrow p.id$ 
                    
                    \STATE $class[ind_p] \leftarrow |vocabulary|$ 
                    \STATE $class[ind_n] \leftarrow |vocabulary|$ 
                \ENDIF
                
            \ENDFOR
        \ENDFOR
    \ENDFOR
    \STATE Update $vocabulary$ to include the new class
\ENDWHILE
\end{algorithmic}
\end{algorithm}
\subsection{Detailed Algorithm}
At the core of our approach that we describe in \cref{alg:MDBPE}, we use two representations for each image, one holding the kind of token (\textit{token class})  covering a pixel, and one that holds the ID describing which unique instance of a token the pixel belongs to (\textit{unique ID}), see \cref{fig:token_unique_class}. When writing out the final sequence, we process from top to bottom, left to right, and only write out the token class once per unique ID, at the anchor position.\\
Initially, the token class is \eg the greyscale value or the VQ-VAE codebook index, and the unique ID is simply a unique number per token. In later steps, \eg for a token combined from two tokens that now spans across two pixels, this would mean both pixels have the same class and the same unique ID (see \cref{fig:token_unique_class}).\\
We iterate two phases of our algorithm:\\
\textbf{Count:} We count the occurrences in a table that maps from tuples consisting of a pair of token classes and a constellation vector to an integer value, initially $0$. We move over each grid point to get a token (initially, a pixel) $p$ of an image, applying a binary mask in both directions for a neighbouring grid point with token $n$ (\ie initially including the pixel to the right and the pixel below as possible elements $n$, see \cref{fig:replacements}). We then identify the anchor points for the tokens of both $p$ and $n$ and compute a vector $v_{pn}$ to identify their constellation. We increment the counter for this specific constellation of tokens, \ie increment the entry $(p, n, v_{pn})$ in the map ("\textit{count}" in \cref{alg:MDBPE}), and mark the combination of unique IDs as used for this specific image ("\textit{used}" in \cref{alg:MDBPE}). This is to avoid counting the same constellation at the same place multiple times. As we check both horizontally and vertically for all pixels, this will cover all constellations of all adjacent tokens.\\
\textbf{Replace:} 
We first identify the entry $(p, n, v_{pn})$ in the map with the highest count. We then iterate over all images and pixels again and act on the occurrences of the chosen entry, \ie with the same classes $p, n$ and the same vector $v_{pn}$. For any occurrence, we identify all pixels belonging to this specific constellation (having the same unique ID). We then re-label (merge) all of the corresponding pixels to have the token class of a new token, and replace the unique ID of $n$ with the unique ID of $p$. After doing so for all images, we go back to counting and repeat until we reach the desired vocabulary size.\\
\textbf{Extract Sequences:} We obtain our condensed sequences by enumerating all tokens from top to bottom, left to right, but only ever output the first occurrence of a unique ID (\ie only output the anchor points).
\begin{figure}[h]
    \centering
    \includegraphics[width=0.7\columnwidth]{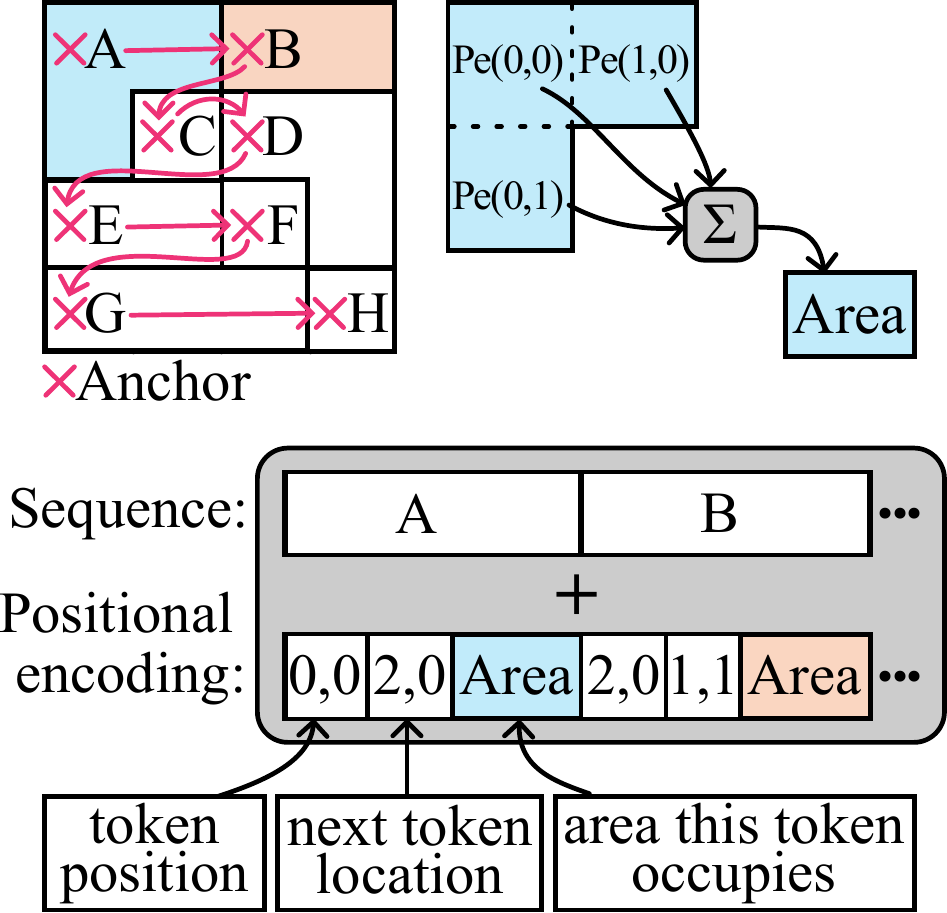}
    \caption{In addition to giving a positional encoding of the token position itself, we also give the network the position of the next token (implicitly defined through previous token shapes) and an integrated positional encoding (\ie sum over $Pe$) describing the area of the token shape.}
    \label{fig:sequence_PE}
\end{figure}

\subsection{Processing Sequences of Varying Length}
We introduce three novel strategies to improve the network's ability to process shortened sequences, beyond the standard positional encoding (see \cref{fig:sequence_PE}):
\paragraph{Next Token Encoding:} While the network implicitly learns the token sizes and therefore, their relative positions, we can also explicitly supply the network with this information to guide the training process. Each token is not only assigned its own position, but also information about the position of the next token. This information is implicitly given by the shape and size of the current and all previous tokens, hence does not interfere with autoregressive training using a causal mask.

\paragraph{Token Shape Encoding:} In addition to encoding the next token’s position, we can also provide the network with information about the spatial coverage of each token. For example, a token occupying a $2$-by-$2$ block should indicate that it covers not only its anchor position, but the remaining three positions as well. To achieve this, we take inspiration from Mip-NeRF \cite{mipnerf}: We integrate over the positional encoding of all positions that a token covers to obtain a description of the token's spatial coverage. In our case of discrete tokens, this integral becomes a sum over the positional encoding of all covered positions:
\[
IPE = \sum_{i \in \text{Token Positions}} \text{positional encoding}(i)
\]

\paragraph{Facilitating Generation:} For autoregressive prediction, we can constrain the output probabilities of tokens based on what is known about the sequence structure: Tokens that would not fit in the sequence (due to exceeding grid boundaries or overlapping with already generated tokens) can have their probabilities clamped to zero or be re-rolled, see \cref{fig:not_generateable}.

\subsection{Boosting Compression Effectiveness}
We can further amplify training on compressed sequences through a lossy formulation: The more tokens we use that are interchangeable without significantly affecting the outcome (different tokens that only signify "blue sky"), the less efficient our compression becomes. Therefore, we propose a lossy approach when obtaining the discrete tokens in the first place: We can combine tokens that have very similar meanings, \ie where replacing one token with another changes the output by only an epsilon. Consider the $1$D example where tokens $A$, $A'$, and $A''$ can be replaced by $A$ without introducing much error in the output. This way, a sequence like $BA'BA''BABABA''$ would become $BABABABABA$, enabling more efficient compression by allowing patterns such as $BA$ to be compressed into a single token.\\
We apply farthest point sampling \cite{qi2017pointnet} on the token embeddings (\eg VQ-VAE \cite{vqvae} or VQGAN \cite{vqgan} codebook), then use this as an initialisation for K-means clustering to obtain $K$ new codebook representatives. Then we snap each codebook index of an image to the closest cluster centre.\\
Training with fewer codewords is not always an option, as is particularly noticeable in VQGANs, which are often used as the backbone for image generation \cite{vqgan, vqgan_alt, vq_video, meta_movie}, as they often fail to converge for smaller sets of codewords. However, collapsing the codebook size in posterior is quite effective, as we can drastically reduce the codebook size (\cref{app:vqgan_dimred}), even without additional decoder fine-tuning on the collapsed codebook.\\
Lastly, we further boost training speed by \textit{pruning} the dataset from the $5\%$ with the longest sequence length: As we pad the sequence length to the largest sequence in the training set for simplicity, this allows us to reduce the sequence length and hence, memory footprint and FLOPs used, drastically. This enables larger batch sizes.
\begin{figure*}[ht]
    \centering
    \begin{minipage}{0.65\textwidth}
        \centering
        \begin{tabular}{|c|c|c|c|c|c|}
			\hline
			Dataset & Any & CIFAR-10 & CIFAR-10 & CIFAR-10 & CIFAR-10 \\
			Compression & None & VQ-VAE & VQ-VAE & VQ-VAE & VQ-VAE \\
			Extra tokens & 0 & 256 & 512 & 1024 & 2048 \\
			\hline
			Compression & 100\% & 92.38\% & 89.16\% & 84.97\% & 79.77\% \\
			\hline
			\hline
			Dataset & CelebA & SVHN & MNIST & ImageNet & ImageNet\\
			Compression & VQ-VAE & VQ-VAE & Greyscale & VQ-VAE & VQGAN\\
			Extra tokens & 512 & 512 & 256 & 512 & 512\\
			\hline
			Compression & 85.89\% & 83.79\% & 54.23\% & 83.74\% & 70\%\dag \\
			\hline
		\end{tabular}
        \caption{Exemplary compression factors across different datasets \cite{imagenet, cifar, mnist, svhn,celeba}, for different numbers of extra tokens. We observe similar behaviour across other datasets, with a tendency of better compression when using larger latent grids. VQ-VAE/VQGAN settings can be found in \cref{app:exact_vq_settings} (Appendix).~~~~~~\textsuperscript{\dag} Achieved with codebook collapse to $128$ codewords.}\label{fig:compressionrates}
    \end{minipage}
    \hfill
    \begin{minipage}{0.30\textwidth}
        \centering
        \includegraphics[width=0.8\linewidth]{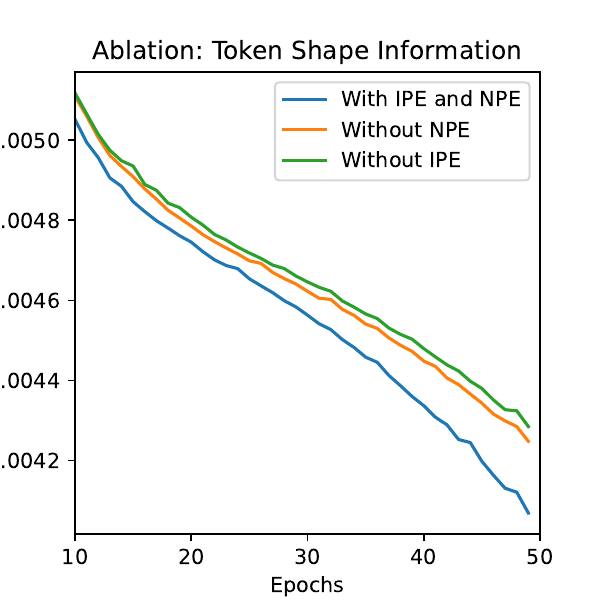}
        \caption{Test loss without positional information for token shape (IPE) and next token position (NPE).}
        \label{fig:pos_enc_impact}
    \end{minipage}
\end{figure*}
\begin{table}[ht]
\centering
\begin{tabular}{|c||c|c||c|c|}
\hline
\multicolumn{1}{|c||}{ Encoding$\rightarrow$} & \multicolumn{2}{c||}{CIFAR-10} & \multicolumn{2}{c|}{CelebA} \\ \hline
\multicolumn{1}{|c||}{Quantisation$\downarrow$} & 1D & MDBPE & 1D & MDBPE \\ \hline
Colours, FID & $68.91$ & \textbf{48.93} & $129.72$ & \textbf{112.05} \\  
Compression & $65\%$ & \textbf{59\%} & $57\%$ & \textbf{54\%}\\ \hline
\hline
VQ-VAE, FID & 36.53 & \textbf{33.93} & 52.15 & \textbf{50.87} \\ 
Compression & $91\%$ & \textbf{89\%} & 87\% & \textbf{86\%} \\ \hline
\end{tabular}
\caption{Comparison of ours (VQ-VAE and MDBPE) to \cite{PLBPE} and in-between mixes on $512$ extra tokens, for generation quality in FID $\downarrow$ after $50$ (CIFAR-10 \cite{cifar}) and $25$ (CelebA \cite{celeba}) epochs. Extended results in \cref{app:plbpe}.}\label{fig:cmp_plbpe}
\end{table}
\begin{figure*}[ht]
    \centering
    \begin{minipage}{0.3\textwidth}
        \centering
        \includegraphics[width=\linewidth]{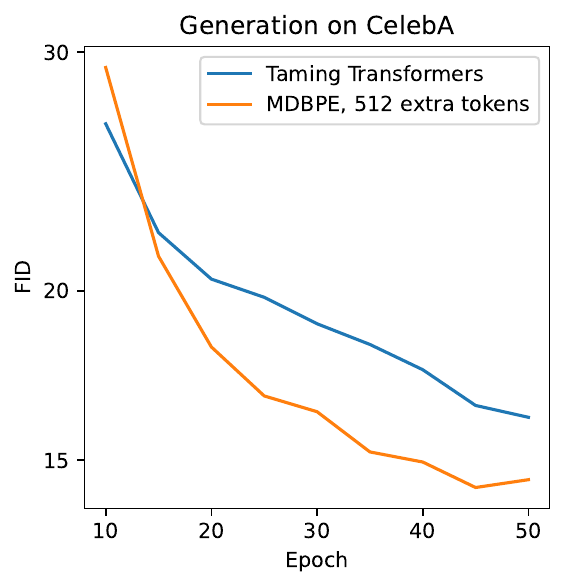}
        \caption{Loss plot comparing to 'Taming Transformers for High-Resolution Image Synthesis' \cite{vqgan} with same conditions.}
        \label{eval:lossplot_vqgan}
    \end{minipage}
    \hfill
    \begin{minipage}{0.35\textwidth}
        \centering
        \captionsetup{type=table}
        \begin{tabular}{|c||c|c|}
            \hline
            Vanilla & With MDBPE & Number of \\
            (in FID) & (in FID) & extra tokens \\
            \hline
            \hline
            \multicolumn{3}{|c|}{MNIST \cite{svhn}, $28^2$, Greyscale at $28^2$}\\
            \hline
            89.07 & \textbf{4.23} & 256  \\
            \hline
            \hline
            \multicolumn{3}{|c|}{CIFAR-10 \cite{cifar} at $32^2$, VQ-VAE at $16^2$}\\
            \hline
            38.32 & \textbf{33.93} & 1024  \\
            \hline
            \hline
            \multicolumn{3}{|c|}{SVHN \cite{svhn} at $32^2$, VQ-VAE at $8^2$}\\
            \hline
            49.79 & \textbf{43.65} & 512  \\
            \hline
            \hline
            \multicolumn{3}{|c|}{CelebA \cite{svhn} at $128^2$, VQ-VAE at $16^2$}\\
            \hline
            65.99 & \textbf{50.87} & 512  \\
            \hline
        \end{tabular}
        \caption{Improved generative performance on different datasets, always given with image- and latent space resolution. Exact specifications in \cref{app:dataset_eval}, scores given in FID $\downarrow$.}\label{tab:diff_datasets}
    \end{minipage}
    \hfill
    \begin{minipage}{0.3\textwidth}
    \centering
        \includegraphics[width=\linewidth]{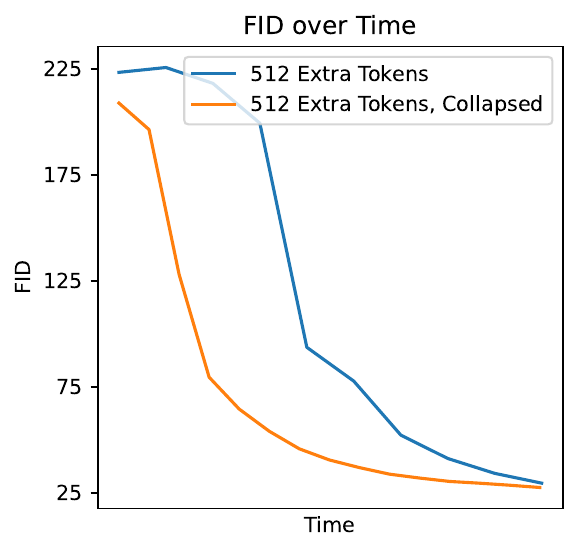}
        \caption{Loss plot comparing lossless variant to lossy (codebook collapse for $K=64$). Collapsed trains for $14$ epochs in the same time lossless trains for $10$ (shorter sequences allow larger batches).}
        \label{eval:lossplot_collapse}
    \end{minipage}
\end{figure*}
\begin{figure*}[htbp]
    \centering
    Training Time $\longrightarrow$\\
    \includegraphics[width=0.8\linewidth]{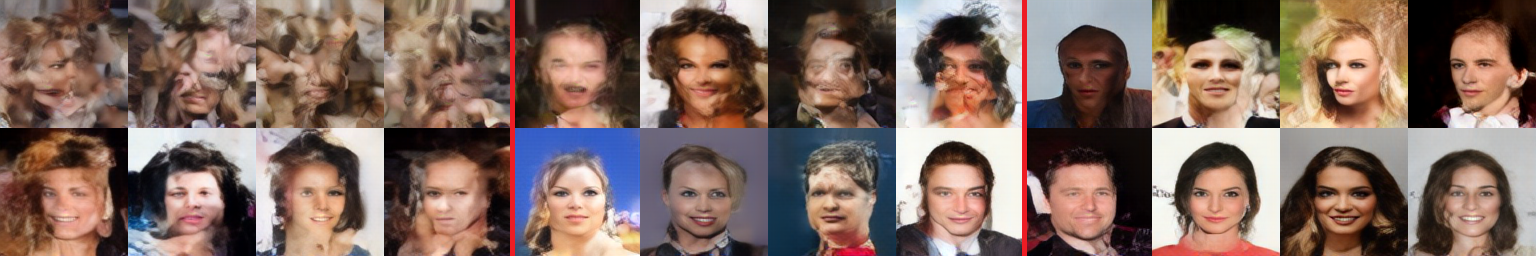}
    \caption{Comparing generation quality of ours with $512$ extra tokens (top) to ours with codebook collapse (bottom), always at about the same time steps (within red lines). This is \textbf{not} the converged result (see \cref{app:diversity}).}
    \label{fig:lossy_vs_ours_none}
\end{figure*}
\begin{figure}
        \captionsetup{type=table}
        \begin{tabular}{|c|c|c|c|c|c|}
            \hline
            \multicolumn{4}{|c|}{Compression} \\
            \hline
            \multicolumn{2}{|c|}{Voxels, $8^3$} & \multicolumn{2}{c|}{Voxels, $32^3$} \\
            \hline
            Cars & Chairs & Planes & Rifles \\
            \hline
            $1.14\%$ & $5.57\%$ & $1.12\%$ & $0.28\%$ \\
            \hline
        \end{tabular}
        \caption{Strong compression rates of our approach on various classes of ShapeNet \cite{shapenet}, using 512 extra tokens.}
        \label{tab:compression_geometry}
\end{figure}
\section{Evaluation}
We choose generation performance expressed through FID \cite{fid} as a measure for most benchmarks, as generating new examples implies a general understanding of the dataset: For both foundation models \cite{radford2018improving, llama3, gpt4} and task-specific approaches \cite{burgess2024viewpointtextualinversiondiscovering, jain2024oneshotlearningmeetsdepth}, training a generative model is the de facto state of the art to learn general information about data \cite{multitask_learners}.\\
Our approach Multidimensional Byte Pair Encoding (MDBPE) is an \textbf{add-on} that is compatible to various representations, on top of any tokenised inputs, \eg a VQ-VAE \cite{vqvae}, VQGAN \cite{vqgan}, greyscale values or quantised colours.\\
We moved most of the exact hyper parameters from training to \cref{app:exact_vq_settings}, following common practice \cite{ViT, vqgan}. All our experiments were run on an Nvidia $2080$ RTX, a consumer graphics card from 2018. We refer to the number of additional tokens in our codebook (vocabulary size increase, not sequence length increase) that our MDBPE produces as \textit{extra tokens}, and refer to the impact on the sequence length as \textit{compression}: A compression of $50 \%$ describes reducing the sequence length by half, always measured on the unseen testset.\\
We ablate hyper parameters (\cref{eval:ablation}) and perform a wide range of multimodal comparisons to show our versatility (\cref{eval:comp}). We investigate the diversity of generated data (\cref{eval:diversity}) and apply MDBPE to the 3D domain (\cref{eval:geometry}). We elaborate the constant complexity compared to BPE in \cref{app:complexity}. Finally, we bring all of our components together to tame transformers further for autoregressive image generation.\\
\textbf{Limitation:} We see the requirement for a previous compression step like VQ-VAE to obtain quantised tokens and the extra processing step as possibly cumbersome. However, both are relatively cheap (here: together much less than $10\%$ of the transformer training time).\\
\textbf{General observations}
\begin{itemize}
    \item \textbf{In training}, tokens covering a larger area can abstract low information density regions (\eg performance goes towards learning difficult tokens rather than optimising the probability of a blue sky token next to another from $99.0$ to $99.9$ percent prediction accuracy).
    \item Shorter sequences mean using drastically fewer FLOPs and faster training due to the quadratic scaling of attention with the number of tokens, while allowing larger batch sizes.
    \item \textbf{In inference}, generating shorter sequences allows less room to make a mistake in the (shorter) sequence that leads to running out of distribution.
    
    \item \textbf{Together}, even moderate compression rates bring significant advantages through both of these factors amplifying each other.
\end{itemize}


\subsection{Ablation}\label{eval:ablation}\label{eval:diversity}
\textbf{VQ-VAE Regularisation}~~~ VQ-VAEs \cite{vqvae} are often regularised to avoid unused codebook entries, \eg through codebook resets or regularisation \cite{wu2020vector, lancucki2020robust, cvqvae, vqwae, svqvae, qgvae}, increasing quality. While using a smaller vocabulary would be highly beneficial for our compression (\ie fewer combinations of tokens mean more repetitions we can compress), we still use a modern VQ-VAE with an additional codebook usage regularisation for all other experiments. Exemplary, on CIFAR-10, this reduces compression effectiveness from $50\%$ to $92\%$ of the sequence length, but allows to eventually reach better FID scores: $36.61$ with regularisation compared to $55.84$ without, measured for $256$ extra tokens after $50$ epochs.\\
\textbf{Number of Extra Tokens}~~~ By design, the first few new tokens introduced by (MD)BPE impact the compression the most. We show the impact of the number of additional tokens in \cref{fig:no_extra_tokens}. The applied transformer has been optimised for the 0-token case and shows that more tokens bring quicker improvements and saturate at better FID values. Through shorter sequence length, \eg with a compression of $70\%$, our models use $30\%$ less tokens to train.\\
\textbf{Modified Positional Encoding}~~~
We show the impact on the loss function of our improved positional encodings in \cref{fig:pos_enc_impact}. The additional token information in the form of area covered by the token and the explicitly given next token position are reducing the loss.\\
\textbf{Different Datasets}~~~ We apply our approach to various image datasets \cite{imagenet, mnist, celeba, svhn} in \cref{tab:diff_datasets} with fixed cut-off dates in training, and show compression rates in \cref{fig:compressionrates}. We observe the transformer to converge faster and make fewer mistakes during generation on shorter sequences, leading to better FID values. We hence interpret the extra spatial structure of our tokens as a helpful feature for additional structure.\\
\textbf{Impact on Diversity}~~~ 
To show that generating shortened sequences does not impact diversity through easier overfitting, we provide two arguments: First, even with strong compression, \eg $2048$ extra tokens on CIFAR-10 and training for $200$ epochs, we observe losses of about $0.0013$ (train) versus $0.0013$ (test) on sequences, showing that we do not overfit. Second, we provide the nearest example from the dataset for CelebA in \cref{app:diversity}.




\subsection{Applications in Geometry}\label{eval:geometry}
\begin{figure*}[ht]
    \centering
    \captionsetup{type=table}
    \begin{minipage}{1.0\textwidth}
        \begin{tabular}{|c|c|c|c|c||c|c|c|c|c|c|c|c|c|}
            \hline
             Generation & \multicolumn{2}{c|}{Cars, $8^3$, Voxels} & \multicolumn{2}{c||}{Chairs, $8^3$, Voxels} & \multicolumn{3}{c|}{Planes, $32^3$, Voxels} & \multicolumn{3}{c|}{Rifles, $32^3$, Voxels} \\
            \hline
            Compression & None & MDBPE & None & MDBPE & None & \cite{octree} & MDBPE & None &\cite{octree} & MDBPE \\
            \hline
            Coverage $\uparrow$ & $58.46\%$ & \textbf{85.13\%} & $35.16\%$ & \textbf{85.20\%} & out of & $47.47\%$ & \textbf{70.47\%} & out of & $69.62\%$ & \textbf{81.52\%}  \\
            MMD $\downarrow$ &  $0.0019$ & \textbf{0.0015} & $0.0097$ & \textbf{0.0049} & VRAM & $0.0031$ & \textbf{0.0019} & VRAM & $0.0024$ & \textbf{0.0016} \\
            \hline
        \end{tabular}
    \end{minipage}
    \caption{Generative results on ShapeNet \cite{shapenet}, with compressed sequences producing more diverse outputs (Coverage) in higher quality (Minimum Matching Distance), as defined in \cite{pmlr-v80-achlioptas18a}. Details in \cref{app:geometry}. We compare to uncompressed sequences for $8^3$ (left half) and to the adaptive 'Octree Transformer' \cite{octree} (right half) for $32^3$, with $512$ extra tokens for MDBPE.}
    \label{tab:generative_geometry}
\end{figure*}
We adapt our approach to geometry by adding a third direction along the z-axis to the pairwise neighbourhoods that we count and replace. With only this minor change, our approach can be applied to \eg binary voxel grids or on the latent codes of a VQ-VAE that produces a cubic grid of tokens representing a Signed Distance Function (SDF).\\ 
Naive transformers, due to their quadratic attention computation, do not apply well to large voxel grids, running out of memory. We give further details and generated examples in \cref{app:geometry}. We show compression rates (\cref{tab:compression_geometry}) and generative performances (\cref{tab:generative_geometry}) both compared to regular transformers on $8^3$ grids and further compare to a state-of-the-art adaptive Octree approach \cite{octree} for large voxel grids in $32^3$.\\
We also apply the principle on tokenised SDFs with 512 extra tokens, compressing the sequence length to $47\%$, with details in \cref{app:geometry}.



\subsection{Comparison to other approaches}\label{eval:comp}
As our approach is applicable to both different domains and is lossless, it is in-between various existing approaches. We hence diversify our comparison: We show that we outperform generation compared to 1D Byte Pair Encoding \cite{PLBPE}, learned adaptive encoding schemes for geometry \cite{octree}, and transformer-based image generation \cite{vqgan}. We show that our approach outperforms all of these approaches, taking less time to reach the same scores and converging to better scores overall.\\
\textbf{Comparison to Pixel-Level BPE}~~~
We advance in three different ways compared to the work of Razzhigaev \etal \cite{PLBPE}: We improve possible compression targets by allowing multidimensional Byte Pair Encoding, improve quality by applying that to a quantised latent space (instead of colour quantisation), and we give the transformer improved positional encoding. We compare results on their chosen datasets, CIFAR-10 \cite{cifar} and CelebA \cite{celeba}, in \cref{fig:cmp_plbpe}, showing better compression and overall generative performance. We show the development over epochs and specify training details in \cref{app:plbpe}.\\
\begin{figure}
    \centering
        \centering
        \includegraphics[width=0.7\linewidth]{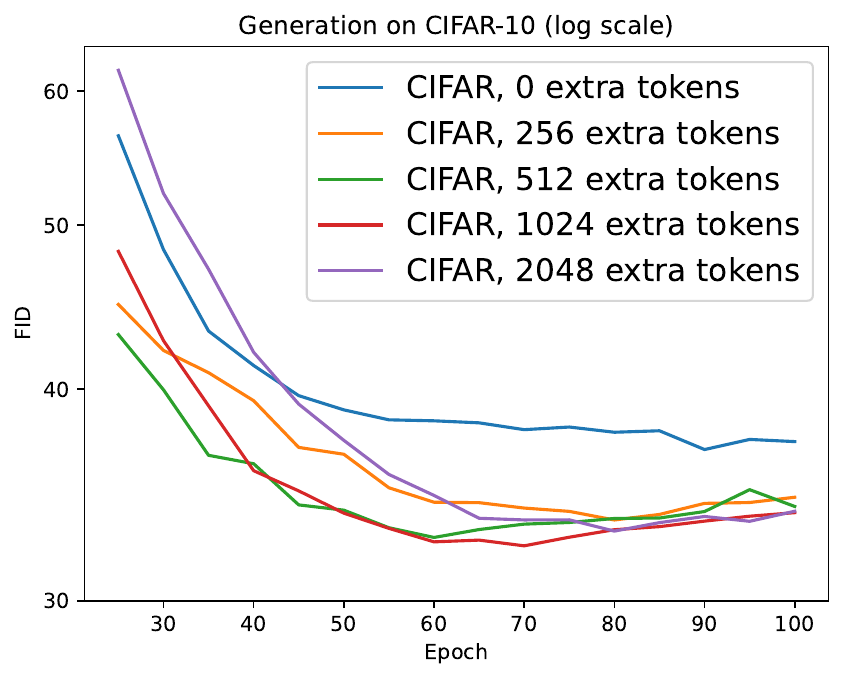}
        \caption{Generation performance boost from introducing extra tokens on CIFAR-10, with an optimum at introducing $1024$ extra tokens.}
    \label{fig:no_extra_tokens}
\end{figure}
\textbf{Octree Transformer}~~~
Octree Transformer \cite{octree} proposes to learn an adaptive resolution representation through an adaptive octree that generates a sequence of codes that indicate whether a cell is further refined. We argue that our method is more flexible by not rigidly splitting a cell evenly, but instead adaptively learning what is good to compress. We compare to their generative performance, trained according to their specifications, in \cref{tab:generative_geometry}. We attribute our stronger results in part to us being able to leverage a higher temperature before outputs collapse: Shorter sequences both offer less room for error and aren't as susceptible when errors occur (\eg erroneously not further refining a section in their sequence can cause much more error than a wrong token for us that only causes a local error).\\
Our approach is simpler, as it has lower overhead and can thus handle batches with about $16$ times as many elements, while each batch (at the same size) already takes only $20$ percent of the time.\\
\textbf{Taming Transformers for High-Resolution Image Synthesis}~~~ To show that our approach is both comparable and competitive, we further compare to and extend 'Taming Transformers for High-Resolution Image Synthesis' \cite{vqgan}. We follow their VQGAN formula, showing that our approach applies well to VQGAN tokens and improves generation performance (\cref{eval:lossplot_vqgan}) compared to their training recipe. Transformers with MDBPE converge to a better optimum, achieving an FID of $13.35$ versus $15.48$ (reached about $20$ epochs later).\\
With our lossy approach based on codebook collapses, we can achieve even better results at a faster pace, taming transformers further. We show the results in \cref{fig:cbcollapse} and provide further details in \cref{app:vqgan_dimred}.\\
\textbf{Further Taming Transformers}~~~Exemplary, for a VQGAN on CelebA, we can increase the effectiveness of our compression by further $20$ percent, which further amplifies generation speed, see  \cref{eval:lossplot_collapse}: Especially for larger or more complex distributions, slightly worse VQGAN performance from collapsing the VQGAN codebook can be very beneficial. We show the FID and compression impact of codebook collapse in \cref{fig:cbcollapse}. We further prune the codebook from the $5\%$ of entries with the longest sequence length, reducing the maximum sequence length to $80\%$. We show results in \cref{fig:lossy_vs_ours_none} and compression rate improvements in \cref{fig:cbcollapse}.
\begin{table}[h]
\centering
\begin{tabular}{|c|c|c|c|c|c|}
\hline
~ & $K=16$ & $K=32$ & $K=64$ & Full \\
\hline
FID$\downarrow$ & 51.19 & 29.78 & 17.85& 7.61 \\
\hline
Compression$\downarrow$ & 49.94\% & 61.75\% & 73.24\% & 100\% \\
\hline
\end{tabular}
\caption{Codebook collapse impact on our VQ-GAN, measured on CelebA\cite{celeba}, $128^2$ pixels with $16^2$ latent size, $10,000$ samples.}\label{fig:cbcollapse}
\end{table}


\section{Conclusion}
Our approach is a simple preprocessing step on top of other compression methods that brings the lossless compression of Byte Pair Encoding to different visual data modalities through leveraging the multidimensional data structure. We show improved performance on condensed sequences through multiple datasets and domains. These compression advantages get further amplified in higher dimensions, \eg 3D data. Our simple, lossy extension of our work enables even further performance gains for \eg rapid prototyping. We see our work as a worthwhile improvement for both prototyping and for researchers with limited hardware. As possible future work, we see further improving positional encodings, \eg by applying 2D versions of RoPE \cite{rope}. Applying different orders of tokens sequences \cite{pixelsnail} and fine-tuning our codebook collapsed VQGAN also seems promising.


{
    \small
    \bibliographystyle{ieeenat_fullname}
    \bibliography{main}

\begin{thebibliography}{69}
\providecommand{\natexlab}[1]{#1}
\providecommand{\url}[1]{\texttt{#1}}
\expandafter\ifx\csname urlstyle\endcsname\relax
  \providecommand{\doi}[1]{doi: #1}\else
  \providecommand{\doi}{doi: \begingroup \urlstyle{rm}\Url}\fi

\bibitem[Achlioptas et~al.(2018)Achlioptas, Diamanti, Mitliagkas, and Guibas]{pmlr-v80-achlioptas18a}
Panos Achlioptas, Olga Diamanti, Ioannis Mitliagkas, and Leonidas Guibas.
\newblock Learning representations and generative models for 3{D} point clouds.
\newblock In \emph{Proceedings of the 35th International Conference on Machine Learning}, pages 40--49. PMLR, 2018.

\bibitem[Aho and Corasick(1975)]{maxmatch1}
Alfred~V Aho and Margaret~J Corasick.
\newblock Efficient string matching: an aid to bibliographic search.
\newblock \emph{Communications of the ACM}, 18\penalty0 (6):\penalty0 333--340, 1975.

\bibitem[Bahdanau et~al.(2014)Bahdanau, Cho, and Bengio]{attention}
Dzmitry Bahdanau, Kyunghyun Cho, and Yoshua Bengio.
\newblock Neural machine translation by jointly learning to align and translate.
\newblock \emph{arXiv preprint arXiv:1409.0473}, 2014.

\bibitem[Bahdanau et~al.(2016)Bahdanau, Cho, and Bengio]{base_attention}
Dzmitry Bahdanau, Kyunghyun Cho, and Yoshua Bengio.
\newblock Neural machine translation by jointly learning to align and translate, 2016.

\bibitem[Barron et~al.(2021)Barron, Mildenhall, Tancik, Hedman, Martin-Brualla, and Srinivasan]{mipnerf}
Jonathan~T Barron, Ben Mildenhall, Matthew Tancik, Peter Hedman, Ricardo Martin-Brualla, and Pratul~P Srinivasan.
\newblock Mip-nerf: A multiscale representation for anti-aliasing neural radiance fields.
\newblock In \emph{Proceedings of the IEEE/CVF international conference on computer vision}, pages 5855--5864, 2021.

\bibitem[Burgess et~al.(2024)Burgess, Wang, and Yeung-Levy]{burgess2024viewpointtextualinversiondiscovering}
James Burgess, Kuan-Chieh Wang, and Serena Yeung-Levy.
\newblock Viewpoint textual inversion: Discovering scene representations and 3d view control in 2d diffusion models, 2024.

\bibitem[Cao et~al.(2023)Cao, Yin, Huang, Liu, Zhao, Zhao, and Huang]{cao2023efficient}
Shiyue Cao, Yueqin Yin, Lianghua Huang, Yu Liu, Xin Zhao, Deli Zhao, and Kaigi Huang.
\newblock Efficient-vqgan: Towards high-resolution image generation with efficient vision transformers.
\newblock In \emph{Proceedings of the IEEE/CVF International Conference on Computer Vision}, pages 7368--7377, 2023.

\bibitem[Chang et~al.(2015)Chang, Funkhouser, Guibas, Hanrahan, Huang, Li, Savarese, Savva, Song, Su, Xiao, Yi, and Yu]{shapenet}
Angel~X. Chang, Thomas~A. Funkhouser, Leonidas~J. Guibas, Pat Hanrahan, Qi{-}Xing Huang, Zimo Li, Silvio Savarese, Manolis Savva, Shuran Song, Hao Su, Jianxiong Xiao, Li Yi, and Fisher Yu.
\newblock Shapenet: An information-rich 3d model repository.
\newblock \emph{CoRR}, abs/1512.03012, 2015.

\bibitem[Chen et~al.(2018)Chen, Mishra, Rohaninejad, and Abbeel]{pixelsnail}
XI Chen, Nikhil Mishra, Mostafa Rohaninejad, and Pieter Abbeel.
\newblock {P}ixel{SNAIL}: An improved autoregressive generative model.
\newblock In \emph{Proceedings of the 35th International Conference on Machine Learning}, pages 864--872. PMLR, 2018.

\bibitem[Cheong et~al.(2023)Cheong, Mustafa, and Gilbert]{cheong2023upgptuniversaldiffusionmodel}
Soon~Yau Cheong, Armin Mustafa, and Andrew Gilbert.
\newblock Upgpt: Universal diffusion model for person image generation, editing and pose transfer, 2023.

\bibitem[Dai et~al.(2023)Dai, Hou, Ma, Tsai, Wang, Wang, Zhang, Vandenhende, Wang, Dubey, et~al.]{dai2023emu}
Xiaoliang Dai, Ji Hou, Chih-Yao Ma, Sam Tsai, Jialiang Wang, Rui Wang, Peizhao Zhang, Simon Vandenhende, Xiaofang Wang, Abhimanyu Dubey, et~al.
\newblock Emu: Enhancing image generation models using photogenic needles in a haystack.
\newblock \emph{arXiv preprint arXiv:2309.15807}, 2023.

\bibitem[Defazio et~al.(2024)Defazio, Xingyu, Yang, Mehta, Mishchenko, Khaled, and Cutkosky]{sfadam}
Aaron Defazio, Xingyu, Yang, Harsh Mehta, Konstantin Mishchenko, Ahmed Khaled, and Ashok Cutkosky.
\newblock The road less scheduled, 2024.

\bibitem[Deiseroth et~al.(2024)Deiseroth, Brack, Schramowski, Kersting, and Weinbach]{tfree}
Bj{\"o}rn Deiseroth, Manuel Brack, Patrick Schramowski, Kristian Kersting, and Samuel Weinbach.
\newblock T-free: Tokenizer-free generative llms via sparse representations for memory-efficient embeddings.
\newblock \emph{arXiv preprint arXiv:2406.19223}, 2024.

\bibitem[Deng et~al.(2009)Deng, Dong, Socher, Li, Li, and Fei-Fei]{imagenet}
Jia Deng, Wei Dong, Richard Socher, Li-Jia Li, Kai Li, and Li Fei-Fei.
\newblock Imagenet: A large-scale hierarchical image database.
\newblock In \emph{2009 IEEE conference on computer vision and pattern recognition}, pages 248--255. Ieee, 2009.

\bibitem[Deng(2012)]{mnist}
Li Deng.
\newblock The mnist database of handwritten digit images for machine learning research.
\newblock \emph{IEEE Signal Processing Magazine}, 29\penalty0 (6):\penalty0 141--142, 2012.

\bibitem[Dosovitskiy et~al.(2020)Dosovitskiy, Beyer, Kolesnikov, Weissenborn, Zhai, Unterthiner, Dehghani, Minderer, Heigold, Gelly, Uszkoreit, and Houlsby]{ViT}
Alexey Dosovitskiy, Lucas Beyer, Alexander Kolesnikov, Dirk Weissenborn, Xiaohua Zhai, Thomas Unterthiner, Mostafa Dehghani, Matthias Minderer, Georg Heigold, Sylvain Gelly, Jakob Uszkoreit, and Neil Houlsby.
\newblock An image is worth 16x16 words: Transformers for image recognition at scale.
\newblock \emph{CoRR}, abs/2010.11929, 2020.

\bibitem[Dubey et~al.(2024)Dubey, Jauhri, Pandey, Kadian, Al-Dahle, Letman, Mathur, Schelten, Yang, Fan, et~al.]{llama3}
Abhimanyu Dubey, Abhinav Jauhri, Abhinav Pandey, Abhishek Kadian, Ahmad Al-Dahle, Aiesha Letman, Akhil Mathur, Alan Schelten, Amy Yang, Angela Fan, et~al.
\newblock The llama 3 herd of models.
\newblock \emph{arXiv preprint arXiv:2407.21783}, 2024.

\bibitem[Elsner et~al.(2024)Elsner, Usinger, Czech, Kobsik, He, Lim, and Kobbelt]{qgvae}
Tim Elsner, Paula Usinger, Victor Czech, Gregor Kobsik, Yanjiang He, Isaak Lim, and Leif Kobbelt.
\newblock Quantised global autoencoder: A holistic approach to representing visual data.
\newblock \emph{arXiv preprint arXiv:2407.11913}, 2024.

\bibitem[Esser et~al.(2020)Esser, Rombach, and Ommer]{vqgan}
Patrick Esser, Robin Rombach, and Bj{\"{o}}rn Ommer.
\newblock Taming transformers for high-resolution image synthesis.
\newblock \emph{CoRR}, abs/2012.09841, 2020.

\bibitem[Feng et~al.(2024)Feng, Weng, Wang, Yuan, Bao, Luo, Chen, and Guo]{feng2024ccedit}
Ruoyu Feng, Wenming Weng, Yanhui Wang, Yuhui Yuan, Jianmin Bao, Chong Luo, Zhibo Chen, and Baining Guo.
\newblock Ccedit: Creative and controllable video editing via diffusion models.
\newblock In \emph{Proceedings of the IEEE/CVF Conference on Computer Vision and Pattern Recognition}, pages 6712--6722, 2024.

\bibitem[Goodfellow et~al.(2014)Goodfellow, Pouget-Abadie, Mirza, Xu, Warde-Farley, Ozair, Courville, and Bengio]{gan}
Ian Goodfellow, Jean Pouget-Abadie, Mehdi Mirza, Bing Xu, David Warde-Farley, Sherjil Ozair, Aaron Courville, and Yoshua Bengio.
\newblock Generative adversarial nets.
\newblock In \emph{Advances in Neural Information Processing Systems}. Curran Associates, Inc., 2014.

\bibitem[Gu et~al.(2022)Gu, Chen, Bao, Wen, Zhang, Chen, Yuan, and Guo]{vec_gen}
Shuyang Gu, Dong Chen, Jianmin Bao, Fang Wen, Bo Zhang, Dongdong Chen, Lu Yuan, and Baining Guo.
\newblock Vector quantized diffusion model for text-to-image synthesis.
\newblock In \emph{Proceedings of the IEEE/CVF conference on computer vision and pattern recognition}, pages 10696--10706, 2022.

\bibitem[Heusel et~al.(2017)Heusel, Ramsauer, Unterthiner, Nessler, Klambauer, and Hochreiter]{fid}
Martin Heusel, Hubert Ramsauer, Thomas Unterthiner, Bernhard Nessler, G{\"{u}}nter Klambauer, and Sepp Hochreiter.
\newblock Gans trained by a two time-scale update rule converge to a nash equilibrium.
\newblock \emph{CoRR}, abs/1706.08500, 2017.

\bibitem[Hinton and Salakhutdinov(2006)]{hinton2006reducing}
Geoffrey~E Hinton and Ruslan~R Salakhutdinov.
\newblock Reducing the dimensionality of data with neural networks.
\newblock \emph{science}, 313\penalty0 (5786):\penalty0 504--507, 2006.

\bibitem[Ho et~al.(2020)Ho, Jain, and Abbeel]{diffusion}
Jonathan Ho, Ajay Jain, and Pieter Abbeel.
\newblock Denoising diffusion probabilistic models.
\newblock In \emph{Advances in Neural Information Processing Systems}, pages 6840--6851. Curran Associates, Inc., 2020.

\bibitem[Huang et~al.()Huang, Mao, Chen, and Zhang]{dqvae}
Mengqi Huang, Zhendong Mao, Zhuowei Chen, and Yongdong Zhang.
\newblock Towards accurate image coding: Improved autoregressive image generation with dynamic vector quantization (supplementary material).

\bibitem[Ibing et~al.(2023)Ibing, Kobsik, and Kobbelt]{octree}
Moritz Ibing, Gregor Kobsik, and Leif Kobbelt.
\newblock Octree transformer: Autoregressive 3d shape generation on hierarchically structured sequences.
\newblock In \emph{Proceedings of the IEEE/CVF Conference on Computer Vision and Pattern Recognition}, pages 2698--2707, 2023.

\bibitem[Issenhuth et~al.(2021)Issenhuth, Tanielian, Mary, and Picard]{edibert}
Thibaut Issenhuth, Ugo Tanielian, J{\'e}r{\'e}mie Mary, and David Picard.
\newblock Edibert, a generative model for image editing.
\newblock \emph{arXiv preprint arXiv:2111.15264}, 2021.

\bibitem[Jain(2024)]{jain2024oneshotlearningmeetsdepth}
Anisha Jain.
\newblock One-shot learning meets depth diffusion in multi-object videos, 2024.

\bibitem[Kaplan et~al.(2020)Kaplan, McCandlish, Henighan, Brown, Chess, Child, Gray, Radford, Wu, and Amodei]{kaplan2020scaling}
Jared Kaplan, Sam McCandlish, Tom Henighan, Tom~B Brown, Benjamin Chess, Rewon Child, Scott Gray, Alec Radford, Jeffrey Wu, and Dario Amodei.
\newblock Scaling laws for neural language models.
\newblock \emph{arXiv preprint arXiv:2001.08361}, 2020.

\bibitem[Koner et~al.(2024)Koner, Jain, Jain, Tresp, and Paul]{koner2024lookupvit}
Rajat Koner, Gagan Jain, Prateek Jain, Volker Tresp, and Sujoy Paul.
\newblock Lookupvit: Compressing visual information to a limited number of tokens.
\newblock \emph{arXiv preprint arXiv:2407.12753}, 2024.

\bibitem[Krizhevsky et~al.(2009)Krizhevsky, Hinton, et~al.]{cifar}
Alex Krizhevsky, Geoffrey Hinton, et~al.
\newblock Learning multiple layers of features from tiny images.
\newblock 2009.

\bibitem[Kudo and Richardson(2018)]{kudo2018sentencepiecesimplelanguageindependent}
Taku Kudo and John Richardson.
\newblock Sentencepiece: A simple and language independent subword tokenizer and detokenizer for neural text processing, 2018.

\bibitem[Köpf et~al.(2023)Köpf, Kilcher, von Rütte, Anagnostidis, Tam, Stevens, Barhoum, Duc, Stanley, Nagyfi, ES, Suri, Glushkov, Dantuluri, Maguire, Schuhmann, Nguyen, and Mattick]{openassistant}
Andreas Köpf, Yannic Kilcher, Dimitri von Rütte, Sotiris Anagnostidis, Zhi-Rui Tam, Keith Stevens, Abdullah Barhoum, Nguyen~Minh Duc, Oliver Stanley, Richárd Nagyfi, Shahul ES, Sameer Suri, David Glushkov, Arnav Dantuluri, Andrew Maguire, Christoph Schuhmann, Huu Nguyen, and Alexander Mattick.
\newblock Openassistant conversations -- democratizing large language model alignment, 2023.

\bibitem[{\L}a{\'n}cucki et~al.(2020){\L}a{\'n}cucki, Chorowski, Sanchez, Marxer, Chen, Dolfing, Khurana, Alum{\"a}e, and Laurent]{lancucki2020robust}
Adrian {\L}a{\'n}cucki, Jan Chorowski, Guillaume Sanchez, Ricard Marxer, Nanxin Chen, Hans~JGA Dolfing, Sameer Khurana, Tanel Alum{\"a}e, and Antoine Laurent.
\newblock Robust training of vector quantized bottleneck models.
\newblock In \emph{2020 International Joint Conference on Neural Networks (IJCNN)}, pages 1--7. IEEE, 2020.

\bibitem[Lee et~al.(2022)Lee, Kim, Kim, Cho, and Han]{vqgan_alt}
Doyup Lee, Chiheon Kim, Saehoon Kim, Minsu Cho, and Wook-Shin Han.
\newblock Autoregressive image generation using residual quantization.
\newblock In \emph{Proceedings of the IEEE/CVF Conference on Computer Vision and Pattern Recognition}, pages 11523--11532, 2022.

\bibitem[Lipman et~al.(2022)Lipman, Chen, Ben-Hamu, Nickel, and Le]{lipman2022flow}
Yaron Lipman, Ricky~TQ Chen, Heli Ben-Hamu, Maximilian Nickel, and Matt Le.
\newblock Flow matching for generative modeling.
\newblock \emph{arXiv preprint arXiv:2210.02747}, 2022.

\bibitem[Liu et~al.(2015)Liu, Luo, Wang, and Tang]{celeba}
Ziwei Liu, Ping Luo, Xiaogang Wang, and Xiaoou Tang.
\newblock Deep learning face attributes in the wild.
\newblock In \emph{Proceedings of International Conference on Computer Vision (ICCV)}, 2015.

\bibitem[Loshchilov and Hutter(2019)]{adamw}
Ilya Loshchilov and Frank Hutter.
\newblock Decoupled weight decay regularization, 2019.

\bibitem[Netzer et~al.(2011)Netzer, Wang, Coates, Bissacco, Wu, Ng, et~al.]{svhn}
Yuval Netzer, Tao Wang, Adam Coates, Alessandro Bissacco, Baolin Wu, Andrew~Y Ng, et~al.
\newblock Reading digits in natural images with unsupervised feature learning.
\newblock In \emph{NIPS workshop on deep learning and unsupervised feature learning}, page~4. Granada, 2011.

\bibitem[OpenAI et~al.(2024)OpenAI, Achiam, Adler, Agarwal, Ahmad, Akkaya, et~al.]{gpt4}
OpenAI, Josh Achiam, Steven Adler, Sandhini Agarwal, Lama Ahmad, Ilge Akkaya, et~al.
\newblock Gpt-4 technical report, 2024.

\bibitem[Palmer(2000)]{maxmatch2}
David~D Palmer.
\newblock Tokenisation and sentence segmentation.
\newblock \emph{Handbook of natural language processing}, pages 11--35, 2000.

\bibitem[Polyak et~al.(2024)Polyak, Zohar, Brown, Tjandra, Sinha, Lee, Vyas, Shi, Ma, Chuang, et~al.]{meta_movie}
Adam Polyak, Amit Zohar, Andrew Brown, Andros Tjandra, Animesh Sinha, Ann Lee, Apoorv Vyas, Bowen Shi, Chih-Yao Ma, Ching-Yao Chuang, et~al.
\newblock Movie gen: A cast of media foundation models.
\newblock \emph{arXiv preprint arXiv:2410.13720}, 2024.

\bibitem[Qi et~al.(2017)Qi, Su, Mo, and Guibas]{qi2017pointnet}
Charles~R Qi, Hao Su, Kaichun Mo, and Leonidas~J Guibas.
\newblock Pointnet: Deep learning on point sets for 3d classification and segmentation.
\newblock In \emph{Proceedings of the IEEE conference on computer vision and pattern recognition}, pages 652--660, 2017.

\bibitem[Radford et~al.(2018)Radford, Narasimhan, Salimans, and Sutskever]{radford2018improving}
Alec Radford, Karthik Narasimhan, Tim Salimans, and Ilya Sutskever.
\newblock Improving language understanding with unsupervised learning.
\newblock 2018.

\bibitem[Radford et~al.(2019)Radford, Wu, Child, Luan, Amodei, Sutskever, et~al.]{multitask_learners}
Alec Radford, Jeffrey Wu, Rewon Child, David Luan, Dario Amodei, Ilya Sutskever, et~al.
\newblock Language models are unsupervised multitask learners.
\newblock \emph{OpenAI blog}, 1\penalty0 (8):\penalty0 9, 2019.

\bibitem[Ramesh et~al.(2021)Ramesh, Pavlov, Goh, Gray, Voss, Radford, Chen, and Sutskever]{dalle}
Aditya Ramesh, Mikhail Pavlov, Gabriel Goh, Scott Gray, Chelsea Voss, Alec Radford, Mark Chen, and Ilya Sutskever.
\newblock Zero-shot text-to-image generation.
\newblock \emph{CoRR}, abs/2102.12092, 2021.

\bibitem[Rampas et~al.(2022)Rampas, Pernias, and Aubreville]{paella}
Dominic Rampas, Pablo Pernias, and Marc Aubreville.
\newblock A novel sampling scheme for text-and image-conditional image synthesis in quantized latent spaces.
\newblock \emph{arXiv preprint arXiv:2211.07292}, 2022.

\bibitem[Razavi et~al.(2019)Razavi, van~den Oord, and Vinyals]{vqvae2}
Ali Razavi, Aaron van~den Oord, and Oriol Vinyals.
\newblock Generating diverse high-fidelity images with vq-vae-2, 2019.

\bibitem[Razzhigaev et~al.(2022)Razzhigaev, Voronov, Kaznacheev, Kuznetsov, Dimitrov, and Panchenko]{PLBPE}
Anton Razzhigaev, Anton Voronov, Andrey Kaznacheev, Andrey Kuznetsov, Denis Dimitrov, and Alexander Panchenko.
\newblock Pixel-level {BPE} for auto-regressive image generation.
\newblock In \emph{Proceedings of the First Workshop on Performance and Interpretability Evaluations of Multimodal, Multipurpose, Massive-Scale Models}, pages 26--30, Virtual, 2022. International Conference on Computational Linguistics.

\bibitem[Ronneberger et~al.(2015)Ronneberger, Fischer, and Brox]{unet}
Olaf Ronneberger, Philipp Fischer, and Thomas Brox.
\newblock U-net: Convolutional networks for biomedical image segmentation.
\newblock \emph{CoRR}, abs/1505.04597, 2015.

\bibitem[Rumelhart et~al.(1986)Rumelhart, Hinton, and Williams]{first_ae}
David~E Rumelhart, Geoffrey~E Hinton, and Ronald~J Williams.
\newblock Learning internal representations by error propagation, parallel distributed processing, explorations in the microstructure of cognition, ed. de rumelhart and j. mcclelland. vol. 1. 1986.
\newblock \emph{Biometrika}, 71\penalty0 (599-607):\penalty0 6, 1986.

\bibitem[Schmidhuber(1992)]{sgan}
Jürgen Schmidhuber.
\newblock Learning factorial codes by predictability minimization.
\newblock \emph{Neural Computation}, 4\penalty0 (6):\penalty0 863--879, 1992.

\bibitem[Sennrich et~al.(2015)Sennrich, Haddow, and Birch]{bpe2}
Rico Sennrich, Barry Haddow, and Alexandra Birch.
\newblock Neural machine translation of rare words with subword units.
\newblock \emph{CoRR}, abs/1508.07909, 2015.

\bibitem[Shi et~al.(2024)Shi, Gu, Xu, Xu, Zhang, and Wang]{shi2024bivdifftrainingfreeframeworkgeneralpurpose}
Fengyuan Shi, Jiaxi Gu, Hang Xu, Songcen Xu, Wei Zhang, and Limin Wang.
\newblock Bivdiff: A training-free framework for general-purpose video synthesis via bridging image and video diffusion models, 2024.

\bibitem[Shibata et~al.(1999)Shibata, Kida, Fukamachi, Takeda, Shinohara, Shinohara, and Arikawa]{bpe}
Yusuxke Shibata, Takuya Kida, Shuichi Fukamachi, Masayuki Takeda, Ayumi Shinohara, Takeshi Shinohara, and Setsuo Arikawa.
\newblock Byte pair encoding: A text compression scheme that accelerates pattern matching.
\newblock 1999.

\bibitem[Song et~al.(2020)Song, Salcianu, Song, Dopson, and Zhou]{wordpiece}
Xinying Song, Alex Salcianu, Yang Song, Dave Dopson, and Denny Zhou.
\newblock Linear-time wordpiece tokenization.
\newblock \emph{CoRR}, abs/2012.15524, 2020.

\bibitem[Strong et~al.(2021)Strong, Rohnke, Bonafonte, {\L}ajszczak, and Wood]{svqvae}
Marek Strong, Jonas Rohnke, Antonio Bonafonte, Mateusz {\L}ajszczak, and Trevor Wood.
\newblock Discrete acoustic space for an efficient sampling in neural text-to-speech.
\newblock \emph{arXiv preprint arXiv:2110.12539}, 2021.

\bibitem[Su et~al.(2021)Su, Lu, Pan, Wen, and Liu]{rope}
Jianlin Su, Yu Lu, Shengfeng Pan, Bo Wen, and Yunfeng Liu.
\newblock Roformer: Enhanced transformer with rotary position embedding.
\newblock \emph{CoRR}, abs/2104.09864, 2021.

\bibitem[Szegedy et~al.(2014)Szegedy, Liu, Jia, Sermanet, Reed, Anguelov, Erhan, Vanhoucke, and Rabinovich]{inception}
Christian Szegedy, Wei Liu, Yangqing Jia, Pierre Sermanet, Scott~E. Reed, Dragomir Anguelov, Dumitru Erhan, Vincent Vanhoucke, and Andrew Rabinovich.
\newblock Going deeper with convolutions.
\newblock \emph{CoRR}, abs/1409.4842, 2014.

\bibitem[van~den Oord et~al.(2017)van~den Oord, Vinyals, and Kavukcuoglu]{vqvae}
A{\"{a}}ron van~den Oord, Oriol Vinyals, and Koray Kavukcuoglu.
\newblock Neural discrete representation learning.
\newblock \emph{CoRR}, abs/1711.00937, 2017.

\bibitem[Vaswani et~al.(2023)Vaswani, Shazeer, Parmar, Uszkoreit, Jones, Gomez, Kaiser, and Polosukhin]{transformer}
Ashish Vaswani, Noam Shazeer, Niki Parmar, Jakob Uszkoreit, Llion Jones, Aidan~N. Gomez, Lukasz Kaiser, and Illia Polosukhin.
\newblock Attention is all you need, 2023.

\bibitem[Vuong et~al.(2023)Vuong, Le, Zhao, Zheng, Harandi, Cai, and Phung]{vqwae}
Tung-Long Vuong, Trung Le, He Zhao, Chuanxia Zheng, Mehrtash Harandi, Jianfei Cai, and Dinh Phung.
\newblock Vector quantized wasserstein auto-encoder, 2023.

\bibitem[Wang et~al.(2024)Wang, Gangavarapu, Yan, and Rush]{wang2024mambabyte}
Junxiong Wang, Tushaar Gangavarapu, Jing~Nathan Yan, and Alexander~M Rush.
\newblock Mambabyte: Token-free selective state space model.
\newblock \emph{arXiv preprint arXiv:2401.13660}, 2024.

\bibitem[Wu and Flierl(2020)]{wu2020vector}
Hanwei Wu and Markus Flierl.
\newblock Vector quantization-based regularization for autoencoders.
\newblock In \emph{Proceedings of the AAAI Conference on Artificial Intelligence}, pages 6380--6387, 2020.

\bibitem[Yan et~al.(2021)Yan, Zhang, Abbeel, and Srinivas]{vq_video}
Wilson Yan, Yunzhi Zhang, Pieter Abbeel, and Aravind Srinivas.
\newblock Videogpt: Video generation using vq-vae and transformers.
\newblock \emph{arXiv preprint arXiv:2104.10157}, 2021.

\bibitem[Yu et~al.(2024)Yu, Weber, Deng, Shen, Cremers, and Chen]{titok}
Qihang Yu, Mark Weber, Xueqing Deng, Xiaohui Shen, Daniel Cremers, and Liang-Chieh Chen.
\newblock An image is worth 32 tokens for reconstruction and generation.
\newblock \emph{arXiv preprint arXiv:2406.07550}, 2024.

\bibitem[Zheng and Vedaldi(2023)]{cvqvae}
Chuanxia Zheng and Andrea Vedaldi.
\newblock Online clustered codebook.
\newblock In \emph{Proceedings of the IEEE/CVF International Conference on Computer Vision}, pages 22798--22807, 2023.

\bibitem[Zhu et~al.(2023)Zhu, Feng, Chen, Bao, Wang, Chen, Yuan, and Hua]{zhu2023designing}
Zixin Zhu, Xuelu Feng, Dongdong Chen, Jianmin Bao, Le Wang, Yinpeng Chen, Lu Yuan, and Gang Hua.
\newblock Designing a better asymmetric vqgan for stablediffusion.
\newblock \emph{arXiv preprint arXiv:2306.04632}, 2023.

\end{thebibliography}
}
\clearpage
\appendix
\newpage


\begin{figure*}[htbp]
    \centering
    \includegraphics[width=0.9\linewidth]{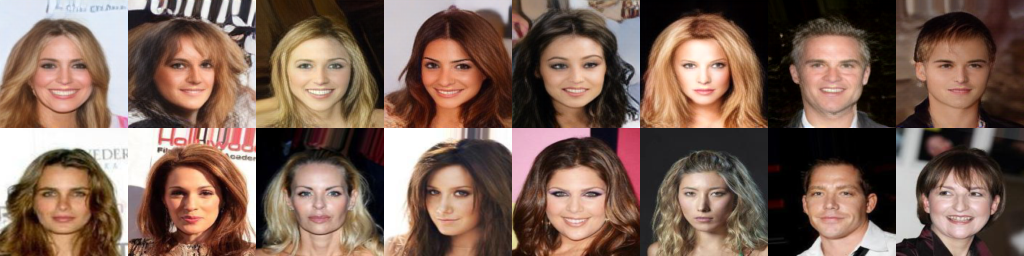}\\
    \caption{Upper row: Generated image, lower row: Closest in Inception\cite{inception} feature vector of the train set.}\label{fig:app_nearest}
\end{figure*}

\section*{Appendix}
\section{Computational Complexity}\label{app:complexity}
The complexity of Byte Pair Encoding is determined by how many pairs of tokens have to be iterated over for counting and replacement. When applying BPE on an image with width and height $W, H$, it can either be flattened into a sequence of length $W \cdot H$, or processed as-is by counting row-wise. This leads to $W \cdot H-1$ pairs to count (flattened) or $(W-1) \cdot H$ (row-wise) possible pairs to count, \ie $\approx W \cdot H$ many pairs. When applying our approach, we count $(W-1) \cdot H$ horizontal pairs and $W \cdot (H-1)$ vertical ones, \ie $\approx 2(W \cdot H)$ pairs to count in total. The same argument holds for the replacement step, hence our overall increase in complexity is just a factor of $2$ compared to BPE.\\
Our (unoptimised) implementation in C++ allows to \eg condense $50,000$ images of $16 \times 16$ tokens by adding $128$ new tokens within less than $4$ minutes on a single core with $3.7$GHz. For large datasets like ImageNet \cite{imagenet} compressed by a VQGAN from $256-by-256$ pixels to $32-by-32$ tokens, a single consumer-grade CPU core can hence increase the vocabulary size by about $25$ tokens per hour. This can be dramatically be reduced by multi CPUs, or only counting on a subset. In relation to training times, the computational overhead is hence neglectable, \ie less than $10\%$ for all our experiments.

\section{Training details}
For all experiments, we use SFAdam \cite{sfadam}, an improved version of Adam \cite{adamw} for transformer optimisation. We always run all experiments ourselves to make sure conditions (like optimiser) are the same, with always using the same foundation if we use \eg a VQ-VAE \cite{vqvae} as base. For training the transformers, we use a learning rate of $0.0002$ instead of $0.001$ when training on compressed sequences for a slight boost in performance (if not specified otherwise). However, our results still hold when using the same learning rate for the examples we tried. We always pad all sequences in a batch to the length of the largest one using end-of-sequence tokens, posing an open angle for future improvements to not waste FLOPs when processing shortened sequences.\\ 
For FID computation on labelled datasets, we use $1000$/$50$ samples pro class (CIFAR-10/MNIST), always comparing with the whole test/validation set; \eg for CIFAR-10, generating $50,0000$ samples to compare to the $10,000$ testset examples would be excessive and only makes a minor difference in the FID scores (the original FID paper\cite{fid} argues to use 50,000 in the light of large datasets ImageNet\cite{imagenet}, which means to use only 500 per class; we hence use double of that for CIFAR/MNIST etc.~ classes). For SVHN and CelebA, we use $10000$ unconditional samples. We use nucleus sampling with a threshold of $0.9$ and no temperature adjustment (temperature $1.0$) for all reported scores, but found the relative difference between runs to be largely invariant to the sampling strategy, \ie while multinomial or topk sampling or adjusted temperature changed the results, the relative improvement between condensed sequence and original sequence stayed largely the same.\\
According to common practise \cite{attention,kaplan2020scaling}, we further always scale our learning rate with $\frac{1}{\sqrt{\frac{N}{54000000}}}$, where $N$ is the number of parameters of our model, as our base model we used for fine-tuning was at 54m parameters.\\
We generally follow the settings of \cite{vqgan} for our encoder-only transformers using the standard pytorch implementation without dropout, split into a small(MNIST, SVHN, CIFAR) and a large sized group (CelebA, Quantised Celeb/CIFAR):
For MNIST: 16 layers, 8 heads, 512 dimensions (54m parameters)\\
For SVHN: 16 layers, 8 heads, 512 dimensions (54m parameters); VQ-VAE with $8^2$, $K=512$\\
For CIFAR: 16 layers, 8 heads, 512 dimensions (54m parameters); VQ-VAE with $16^2$, $K=512$\\
For CelebA (VQ-VAE): 24 layers, 16 heads, 1024 dimensions (211m parameters); VQ-VAE with $8^2$, $K=512$, $128^2$ pixels\\
For CelebA (VQGAN): 24 layers, 16 heads, 1024 dimensions (211m parameters); VQGAN with $16^2$, $K=2048$, $128^2$ pixels\\
For Quantised Celeb/CIFAR: 24 layers, 16 heads, 1024 dimensions (211m params); $32^2$ pixels\\

\section{Comparison to Pixel-Level BPE for Auto-Regressive Image Generation}\label{app:plbpe}
\begin{figure}[htbp]
    \centering
    \includegraphics[width=0.9\linewidth]{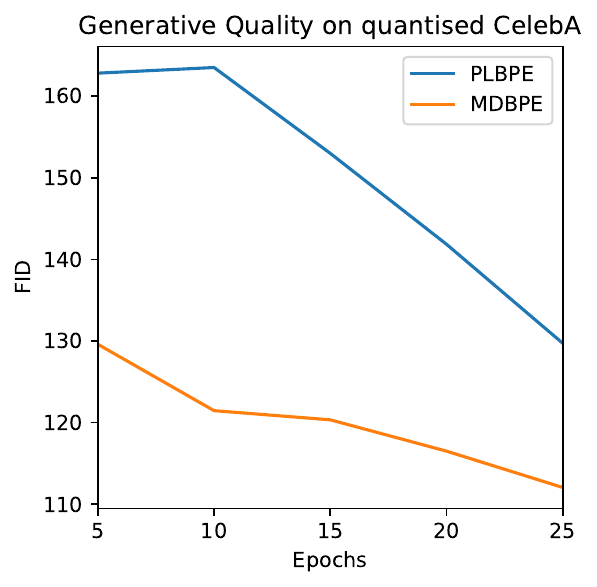} \\
    
    \begin{minipage}{0.05\textwidth}
        \centering
        \rotatebox{90}{\small$\leftarrow$Epochs}
    \end{minipage}%
    \begin{minipage}{0.95\textwidth}
        \includegraphics[width=0.45\linewidth]{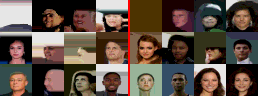}
        \par\medskip
        \small{PLBPE (1D BPE) ~~~~~~~~~~~~~~~~ MDBPE}
    \end{minipage}
    \caption{Generation over time on quantised CelebA at 32-by-32 pixels, comparing PLBPE \cite{PLBPE} to our approach MDBPE, including improved embeddings. Examples show epochs $5, 10, 25$.}\label{fig:app_qceleb}
\end{figure}

\begin{figure}[htbp]
    \centering
    \includegraphics[width=0.9\linewidth]{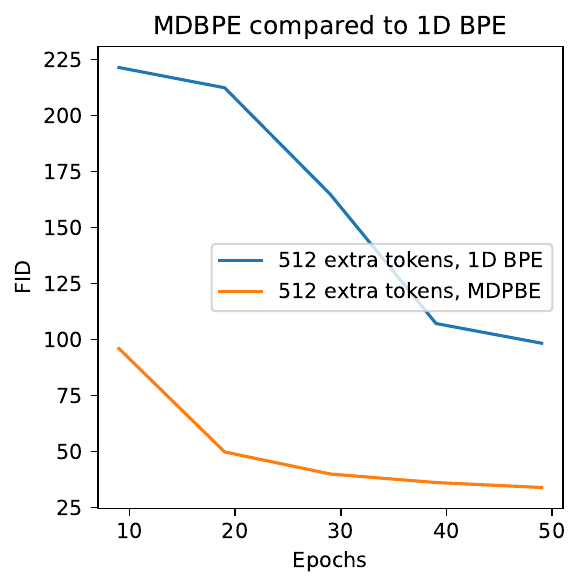} \\
    \caption{Development of FID values when applying MDBPE and the idea behind \cite{PLBPE}, namely, 1D BPE, on CIFAR-10 \cite{cifar}.}\label{fig:app_cifar_ours_vs_1d}
\end{figure}

For colour quantisation, we follow their idea: We use a resolution of $32 \times 32$ pixels while discretising each colour channel to 10 values per R/G/B by dividing by 26, \ie turning an RGB value of $255, 200, 146$ into $9, 7, 5 = 975$. We show the development of the FID values for quantised CelebA \cite{celeba}, together with examples, in \cref{fig:app_qceleb}. Note that we only use $32 \times 32$ pixels here, as we only have Nvidia 2080 TI for our evaluation, and larger images do not fit the memory without compression through a VQ-VAE.\\
We further provide the FID development on CIFAR-10, with a VQ-VAE instead of colour quantisation, in \cref{fig:app_cifar_ours_vs_1d}.

\section{Experiment Specifications on Different Datasets}\label{app:dataset_eval}
For all examples given in \cref{tab:diff_datasets}, we observe similar curves to \cref{fig:no_extra_tokens}, \ie earlier saturation for shortened sequences and better overall performance. For SVHN \cite{svhn} and MNIST \cite{mnist}, we trained for 30 epochs, while for the more complex CIFAR \cite{cifar}, we trained for 50 epochs. We trained 25 epochs for the relatively large CelebA \cite{celeba}.\\

\section{Details on Generating Geometry}\label{app:geometry}
\begin{figure}[htbp]
    \centering
    \includegraphics[width=0.9\linewidth]{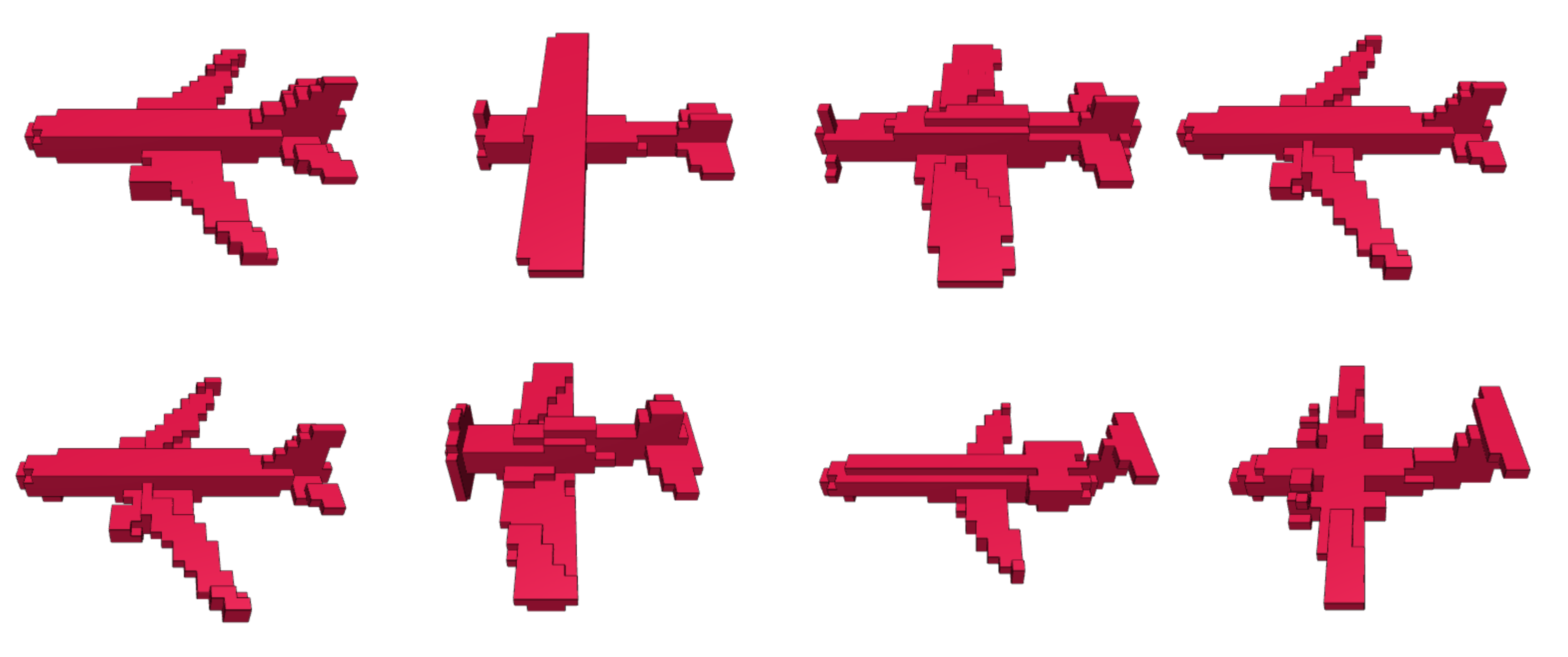} \\
    \includegraphics[width=0.9\linewidth]{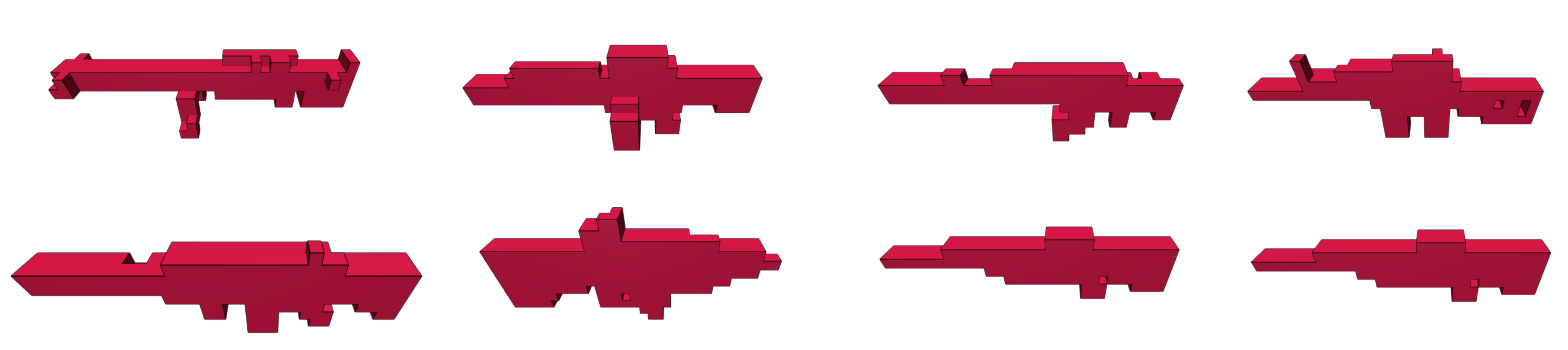} \\
    \includegraphics[width=0.9\linewidth]{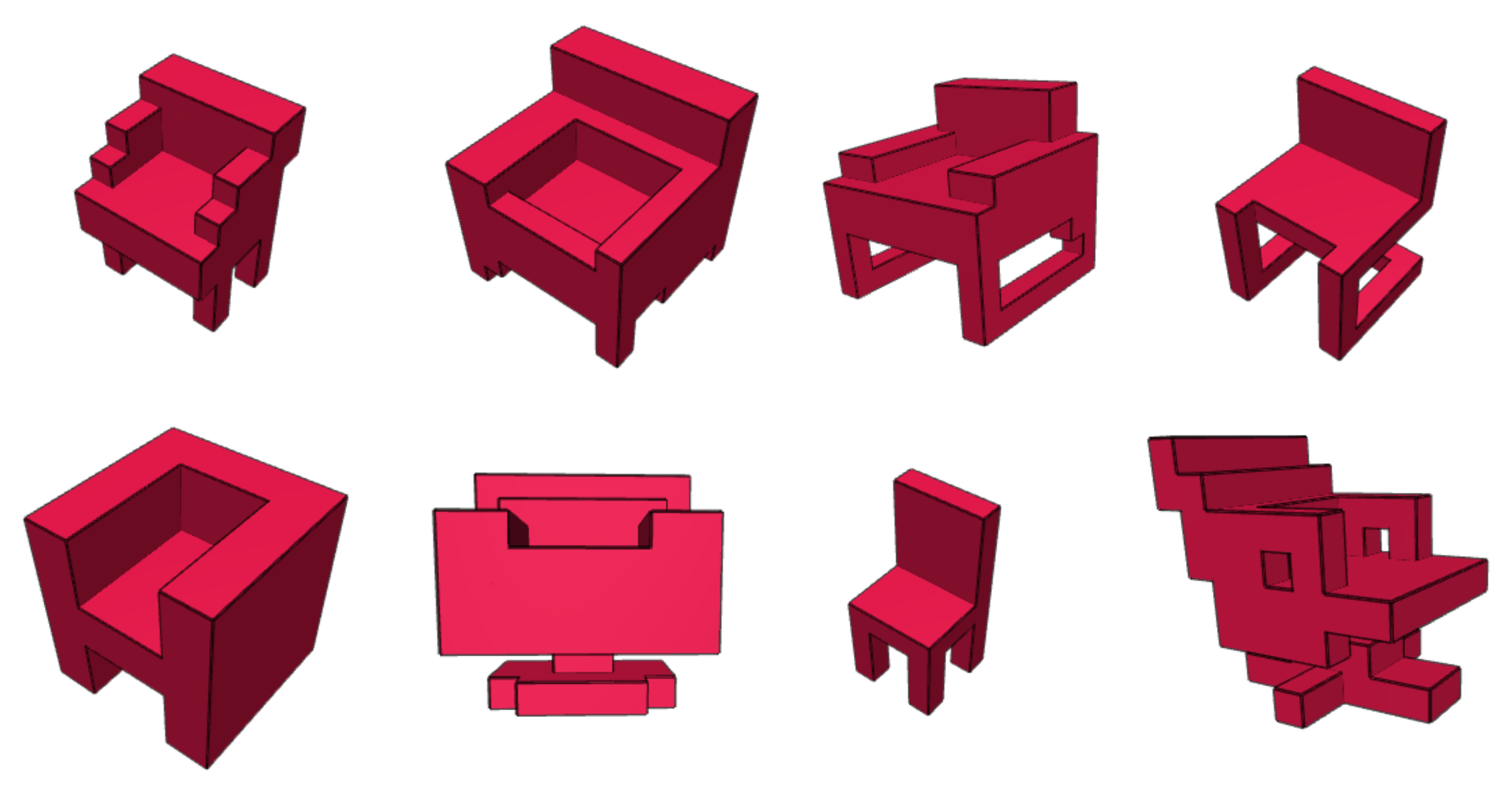} \\
    \caption{Generated voxel grids from ShapeNet \cite{shapenet} at a resolution of $32^3$ (planes, rifles) and $8^3$ (chairs) with our approach. Without compression, attention computation on 32768 voxels for $32^3$ would far exceed the memory of consumer GPUs.}\label{fig:app_generated_results}
\end{figure}

For geometry, we train the transformer for 500 epochs, as the dataset is much smaller, and use weight decay of $0.01$. We use a transformer with $~8$m parameters each, with $256$ dimensions, 8 layers, and 8 heads. For sampling, we use multinomial sampling with temperature $2.0$ (the vanilla transformer would collapse to very few shapes without this adjustment).\\
For MMD and Coverage, we follow\cite{pmlr-v80-achlioptas18a}: We split the data in train and test, with 20 percent of the data as test split, then generate as many examples are in the train split. We then computing coverage by finding the closest example for each generated object in the testset, then computing the percentage of the testset that is "covered" by an example as the nearest, measuring the fidelity of the distribution / the diversity. For MMD, we compute the distance of each testset object to the closest generated one, measuring the faithfulness / the quality of the generated objects. As distance measure, we use chamfer distance on uniform random sampled point clouds of the extracted meshes from surface, with $2048$ samples each. We always sample with a temperature of $1.75$ for $32^2$, as is optimal for the $0$-extra-token-case, for comparability. \\
We show generated examples of $8 \times 8 \times 8$ and $32 \times 32 \times 32$ in \cref{fig:app_generated_results}. We did not use the same classes everywhere for more variation and because \eg the often very thin rifle park do not make much sense in $8^3$ .\\
\paragraph{SDFs} We applied a naive VQ-VAE\cite{vqvae} on a $32^3$ distance field where we clamped the distance values to a fixed value for every cell further than two away from the surface (only voxels with the surface matter for the resulting quality), then apply our compression scheme with $512$ extra tokens on the resulting $8^3$ token grid where it achieved an average compression of $47\%$. We then generated these sequences as usual. Note that our results are not published here, we only add this as a further demonstration of versatility and are not experts on SDFs, \eg the VQ-VAE has a lot of potential. Also, our SDF converter to obtain the SDFs failed on some of the chairs used, hence our scores would not be fair. We always use learning rate $0.00005$ with temperature $1.5$, and (due to the more rich information) use a bigger transformer with 16 layers, 1024 dimensions, and 16 heads.

\section{VQ-VAE Settings for Compression}\label{app:exact_vq_settings}
We apply a VQ-VAE with latent size $8^2$, $K=512$ for SVHN, a VQ-VAE with $16^2$, $K=512$ for CIFAR-10 at $32^2$ pixels and CelebA at $128^2$ pixels, and a VQGAN for CelebA with latent size $16^2$, $K=2048$, and $128^2$ pixels. Note that we always use the same VQ-VAE/VQGAN as baseline for all our experiments for comparability, \ie all gains in performance stem only from our sequence shortening. For SDFs, we use a VQ-VAE with $K=512$ that reduces a $32^3$ latent grid to $8^3$.

\section{Diversity in Shortend Sequences}\label{app:diversity}
We follow the approach of Esser \etal \cite{vqgan} and find the nearest image in the sense of euclidean distance in the feature vectors of the images from the second last layer of the Inception network\cite{inception} from the train dataset for randomly picked generated images in \cref{fig:app_nearest}.

\begin{figure}
    \centering
    \includegraphics[width=0.9\linewidth]{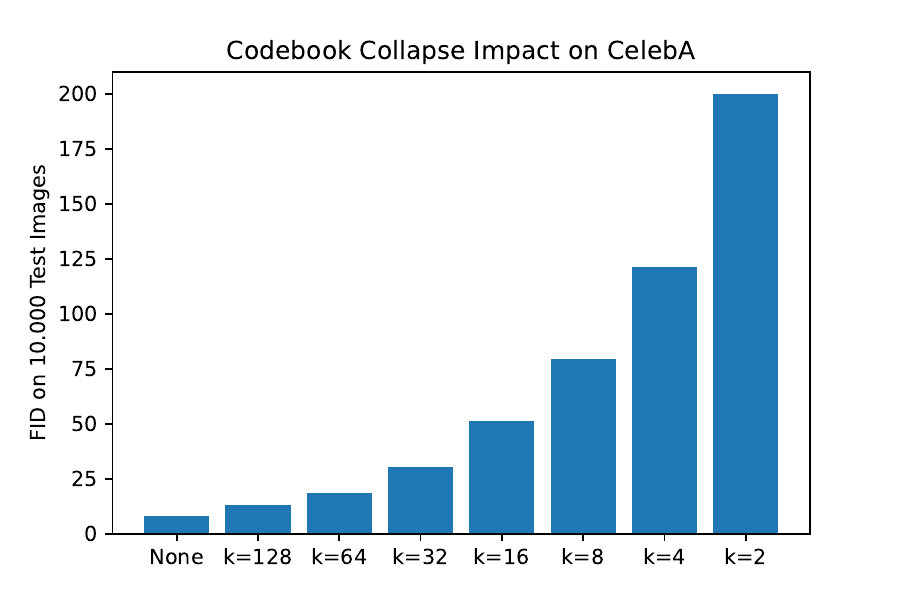}
    \caption{Impact of the number of tokens used, \ie how the codebook collapse barely impacts quality.}\label{fig:no_tokens_used}
\end{figure}

\section{Reducing Tokens in VQGANs}\label{app:vqgan_dimred}
\begin{figure*}
    \centering
    \begin{minipage}{0.05\textwidth}
        \centering
        \rotatebox{90}{Input}
    \end{minipage}%
    \begin{minipage}{0.95\textwidth}
        \centering
        \includegraphics[width=0.65\linewidth]{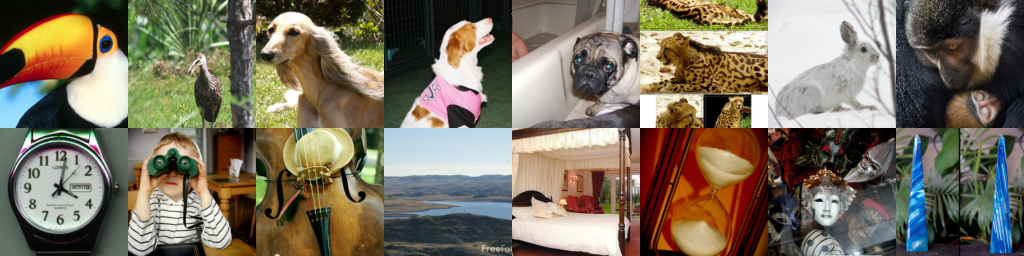}
    \end{minipage}
    \begin{minipage}{0.05\textwidth}
        \centering
        \rotatebox{90}{$K=128$}
    \end{minipage}%
    \begin{minipage}{0.95\textwidth}
        \centering
        \includegraphics[width=0.65\linewidth]{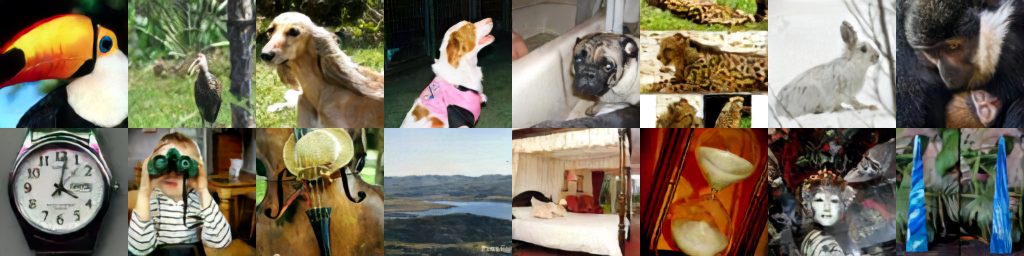}
    \end{minipage}
    \begin{minipage}{0.05\textwidth}
        \centering
        \rotatebox{90}{$K=64$}
    \end{minipage}%
    \begin{minipage}{0.95\textwidth}
        \centering
        \includegraphics[width=0.65\linewidth]{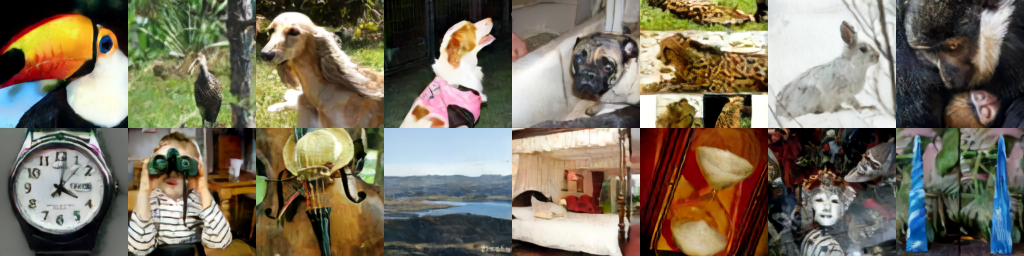}
    \end{minipage}
    \begin{minipage}{0.05\textwidth}
        \centering
        \rotatebox{90}{$K=32$}
    \end{minipage}%
    \begin{minipage}{0.95\textwidth}
        \centering
        \includegraphics[width=0.65\linewidth]{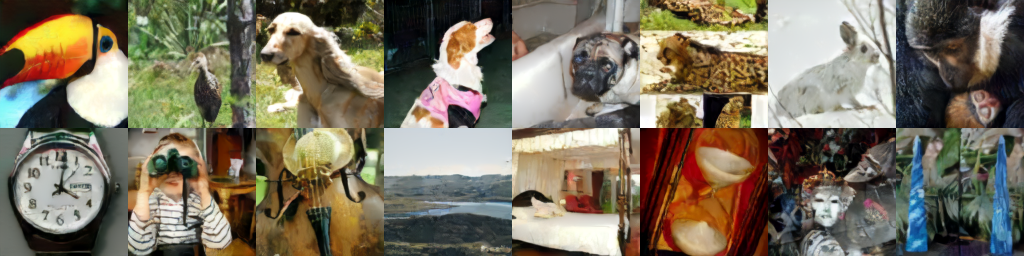}
    \end{minipage}
    \begin{minipage}{0.05\textwidth}
        \centering
        \rotatebox{90}{$K=16$}
    \end{minipage}%
    \begin{minipage}{0.95\textwidth}
        \centering
        \includegraphics[width=0.65\linewidth]{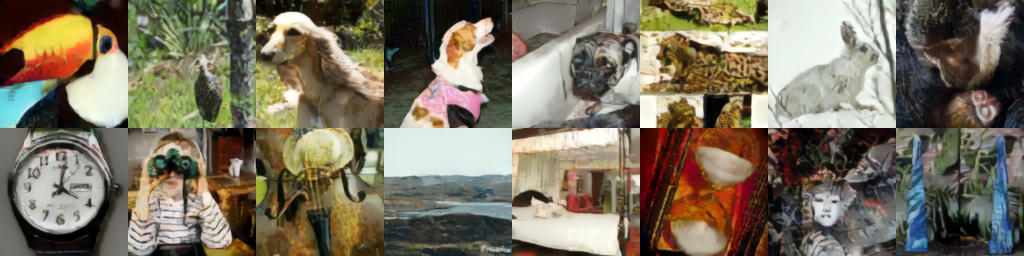}
    \end{minipage}
    \begin{minipage}{0.05\textwidth}
        \centering
        \rotatebox{90}{$K=8$}
    \end{minipage}%
    \begin{minipage}{0.95\textwidth}
        \centering
        \includegraphics[width=0.65\linewidth]{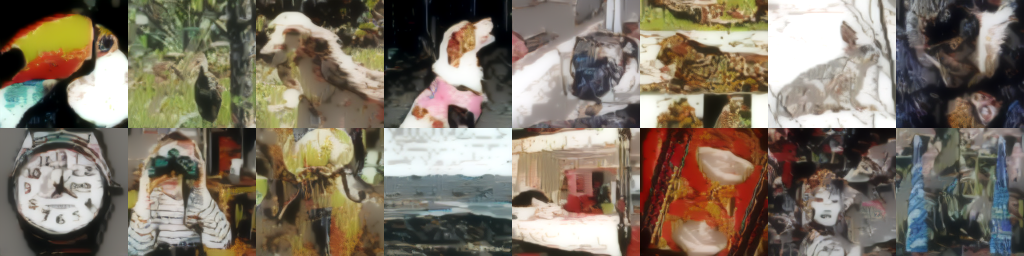}
    \end{minipage}
    \begin{minipage}{0.05\textwidth}
        \centering
        \rotatebox{90}{$K=2$}
    \end{minipage}%
    \begin{minipage}{0.95\textwidth}
        \centering
        \includegraphics[width=0.65\linewidth]{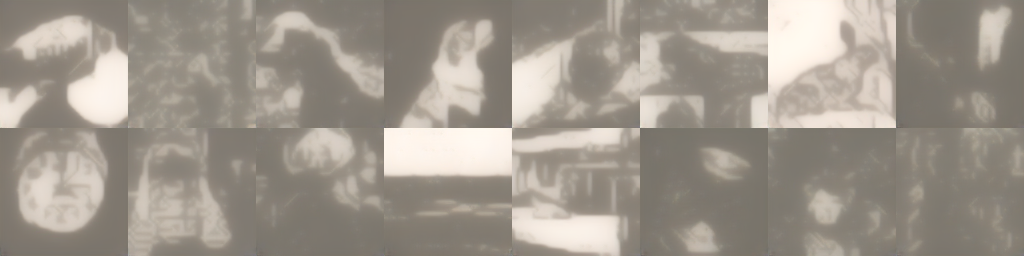}
    \end{minipage}
    \caption{Different numbers of codebook entries $K$ and their impact on the resulting image quality produced by a VQGAN \cite{vqgan} with initially $2048$ codebook entries. Note how dropping from $2048$ to $128$ tokens barely makes a visible difference.}\label{fig:app_vqgan_collapse_imagenet}
\end{figure*}

We evaluate our proposed VQGAN collapse in two different ways: We show qualitatively how the choice of $K$ barely impacts the result in \cref{fig:app_vqgan_collapse_imagenet}, examplary on ImageNet as a more general dataset \cite{imagenet}. We then measure the quantitative impact of the number of tokens used for images of human faces of CelebA \cite{celeba} in \cref{fig:no_tokens_used}: Using only $128$ instead of the full $2048$ codebook entries can produce almost the same fidelity. We mostly observe small changes in the high frequent details of an image for (relatively) high choices of $K$, \eg small earrings appearing and disappearing, that barely affect quality. We also argue that a more powerful VQGAN (ours only uses $28$m parameters, which is orders of magnitude smaller than the state of the art) can probably make up for this even better, and that further fine-tuning on the results would also be beneficial. We leave this for future work, as our focus is on compression, not on tuning VQGANs (and the Nvidia 2080 TI our lab is equipped with barely fit a batch of high resolution images already when only using 28m parameters).

\end{document}